\documentclass[10pt,twocolumn,letterpaper]{article}
\usepackage{algorithm}
\usepackage{algorithmic}
\usepackage{booktabs}
\usepackage{multirow}
\usepackage{graphicx}
\usepackage[pagenumbers]{iccv} % To force page numbers, e.g. for an arXiv version

\definecolor{iccvblue}{rgb}{0.21,0.49,0.74}
\usepackage[pagebackref,breaklinks,colorlinks,allcolors=iccvblue]{hyperref}

\title{Multi-perspective Contrastive Logit Distillation}

\author{Qi Wang\\
Hosei University\\
{\tt\small qi.wang.7c@stu.hosei.ac.jp}
\and
Jinjia Zhou\\
Hosei University\\
{\tt\small zhou@hosei.ac.jp}
}

\begin{document}
\maketitle
\begin{abstract}
In previous studies on knowledge distillation, the significance of logit distillation has frequently been overlooked. To revitalize logit distillation, we present a novel perspective by reconsidering its computation based on the semantic properties of logits and exploring how to utilize it more efficiently. Logits often contain a substantial amount of high-level semantic information; however, the conventional approach of employing logits to compute Kullback-Leibler (KL) divergence does not account for their semantic properties. Furthermore, this direct KL divergence computation fails to fully exploit the potential of logits. To address these challenges, we introduce a novel and efficient logit distillation method, Multi-perspective Contrastive Logit Distillation (MCLD), which substantially improves the performance and efficacy of logit distillation. In comparison to existing logit distillation methods and complex feature distillation methods, MCLD attains state-of-the-art performance in image classification, and transfer learning tasks across multiple datasets, including CIFAR-100, ImageNet, Tiny-ImageNet, and STL-10. Additionally, MCLD exhibits superior training efficiency and outstanding performance with distilling on Vision Transformers, further emphasizing its notable advantages. This study unveils the vast potential of logits in knowledge distillation and seeks to offer valuable insights for future research.
\end{abstract}    
\section{Introduction}
\label{sec:intro}
Over the past decade, remarkable advancements in computer vision have facilitated significant progress across a broad range of tasks, including image classification \cite{vgg,resnet}, semantic segmentation \cite{unet,dconv}, and object detection \cite{fasterrcnn,yolo}.\par
However, due to the increasing demands of real-world tasks, the parameter scale and computational complexity of deep neural networks (DNNs) have grown substantially, resulting in significant memory consumption and computational costs. This challenge poses significant difficulties in deploying neural networks on resource-constrained devices. To address these challenges, various optimization techniques have been devised to accelerate model training and compress neural networks, including lightweight architectures \cite{depthwise,mobilenet}, model quantization \cite{quantization1,quantization2}, model pruning \cite{pruning1,pruning2}, and knowledge distillation \cite{kd,crd,dkd,dot}.\par
\begin{figure}[htbp]
	\centering
	\includegraphics[width=\columnwidth]{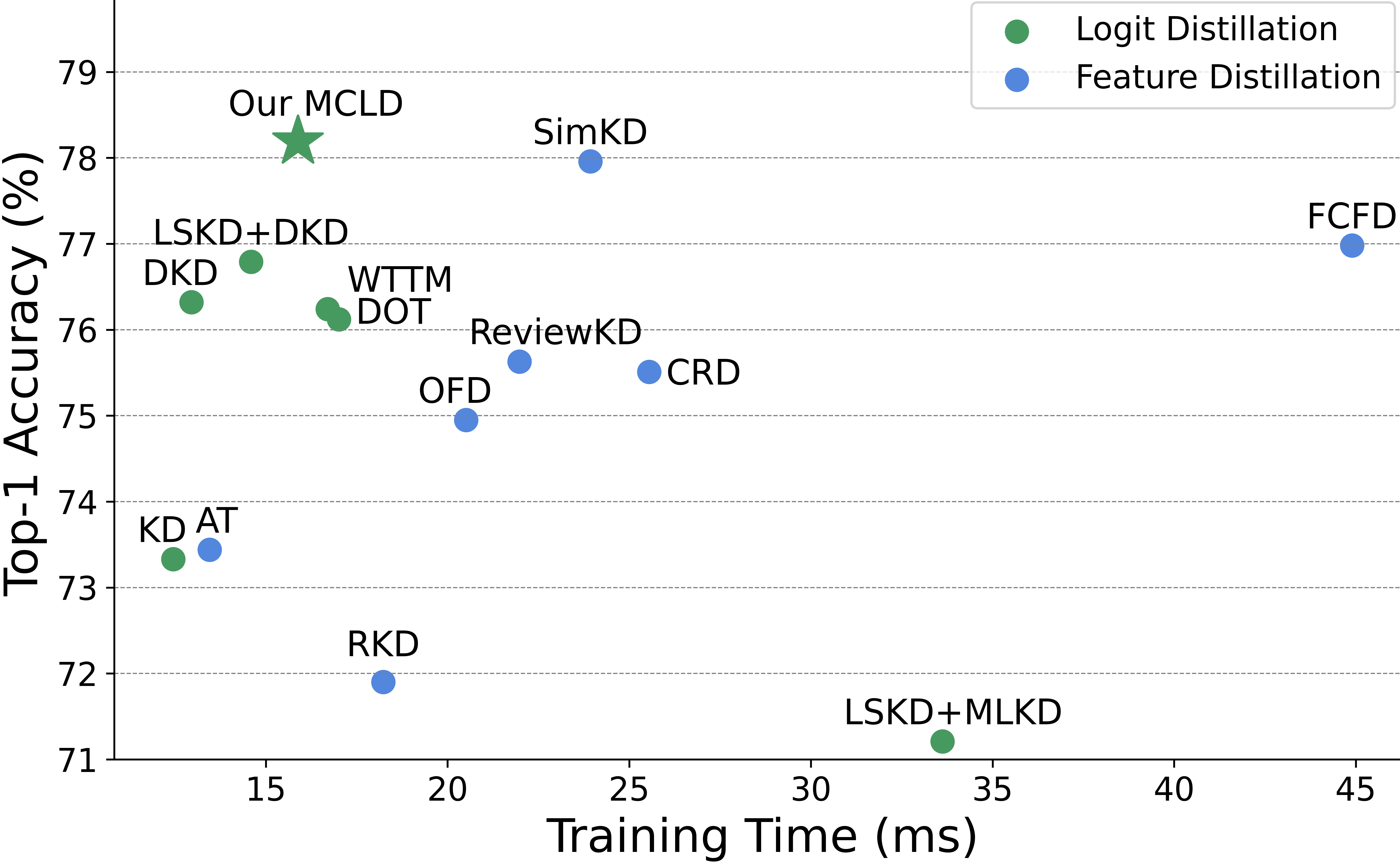}
	\caption{Training time (per batch) vs. Top-1 Accuracy on CIFAR-100 \cite{cifar100}. We set ResNet32×4 as the teacher and ResNet8x4 as the student.}
	\label{fig:kd-comparison}
\end{figure}
In 2015, Geoffrey Hinton et al.\cite{kd} introduced the concept of Knowledge Distillation (KD), which has since been widely adopted for its practicality, efficiency, and versatility. KD can be applied to various network architectures and integrated with other compression techniques, such as pruning and quantization, to further reduce model size. Following \cite{kd}, research on knowledge distillation has proliferated and is typically classified into two main categories (see \cref{fig:kd-comparison}): (1) Logit Distillation \cite{logit3,dkd,dot,lskd,TTM}, which is relatively straightforward to implement and train but performs worse than feature distillation methods due to the lack of information diversity; (2) Feature Distillation \cite{ab,similarity,crd,review,FCFD} contains richer feature information. Compared to logit distillation methods, feature distillation generally achieves superior performance across various tasks but often requires additional modules to align intermediate feature dimensions, leading to increased storage and computational overhead.\par
To overcome these challenges, we propose a novel and efficient logit distillation method, Multi-perspective Contrastive Logit Distillation (MCLD) achieves a trade-off between state-of-the-art performance and training time (As shown in \cref{fig:kd-comparison}).
\subsection*{Main Contributions}
\begin{itemize}
	\item We reconsider the computational mechanism of logit distillation and its properties while identifying ways to utilize logits more efficiently.
	\item Efficiently utilizing logits from multiple perspectives mitigates the lack of diversity in logits.
	\item We introduce a novel logit distillation method, Multi-perspective Contrastive Logit Distillation (MCLD), to overcome the limitations of existing approaches. MCLD attains state-of-the-art performance across diverse tasks and datasets. Furthermore, we empirically demonstrate that MCLD provides higher training efficiency and superior feature transferability compared to feature distillation methods.
\end{itemize}
\section{Related Work}
The concept of Knowledge Distillation (KD), introduced by Hinton et al. \cite{kd}, refers to a learning paradigm where a larger teacher network supervises the training of a smaller student network across multiple tasks \cite{kd,kdobjdetec,kdpose}. Research on KD has since progressed in two primary directions: logit distillation \cite{kd,logit3,dkd,dot} and feature distillation \cite{factor,crd,review,reuse}.\par
Earlier studies \cite{logit3} on logit distillation primarily aimed at developing effective regularization and optimization strategies rather than proposing entirely new methodologies. Additionally, several studies have explored interpretations of classical KD methods \cite{kdquntify,dkd,TTM}. In feature distillation, prior research \cite{factor,ab,review,reuse} has demonstrated excellent performance by transferring the information of feature maps between teacher and student networks, while \cite{crd} introduces a method that integrates contrastive learning with feature distillation. Most feature distillation methods \cite{review,FCFD} achieve superior performance but also incur significant computational and storage costs.
\section{Motivation}
\label{sec:motivation}
Intuitively, logit distillation should be at least comparable to, if not superior to, feature distillation, as logits contain richer high-level semantic information than intermediate features. This insight motivates us to reconsider the potential of logits and the methodology for computing them in logit distillation.\par
\begin{figure}[htbp]
	\centering
	\begin{subfigure}{0.15\textwidth}
		\includegraphics[width=\textwidth]{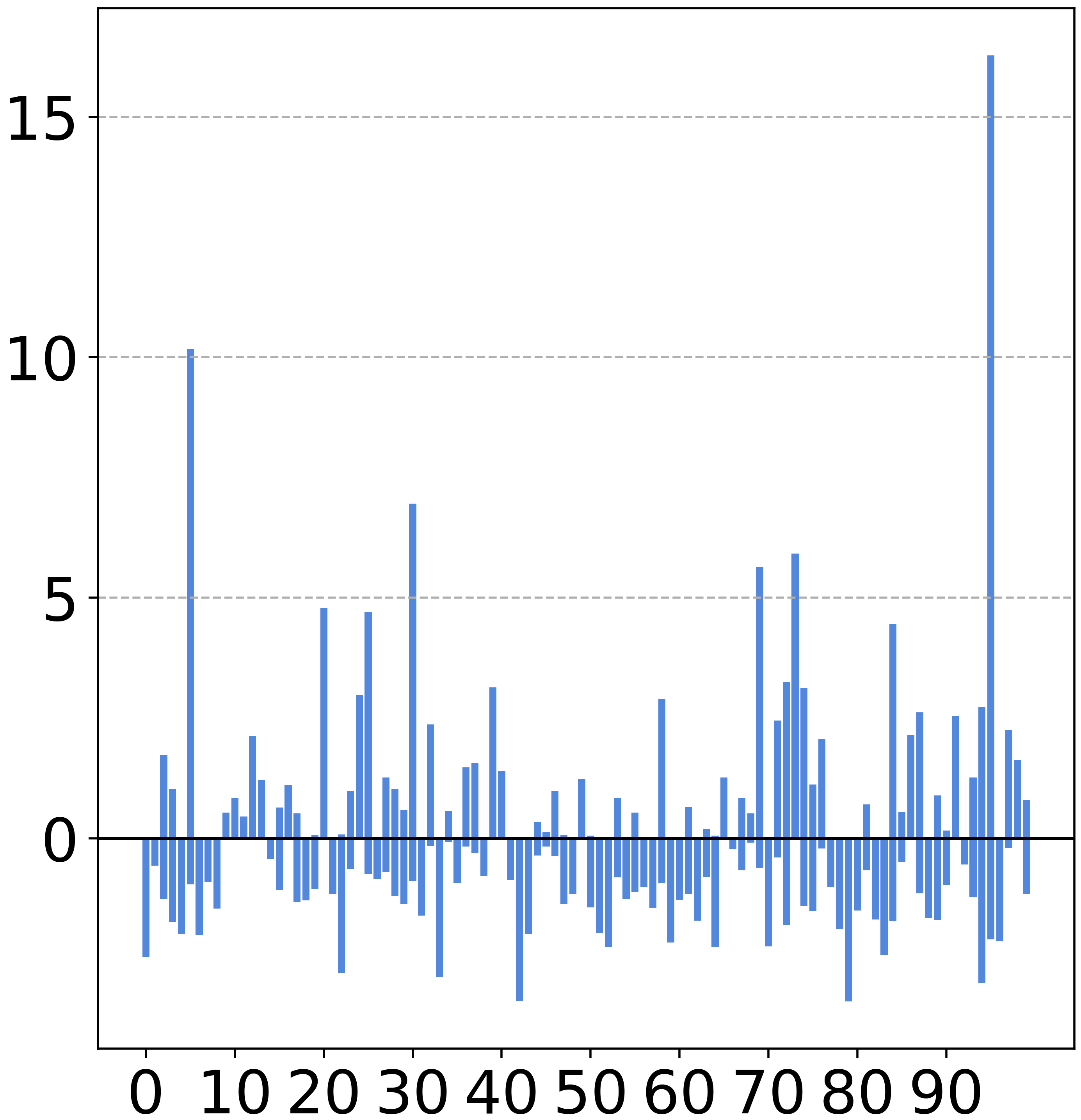}
		\caption{Logits}
		\label{fig:Logits}
	\end{subfigure}
	\begin{subfigure}{0.152\textwidth}
		\includegraphics[width=\textwidth]{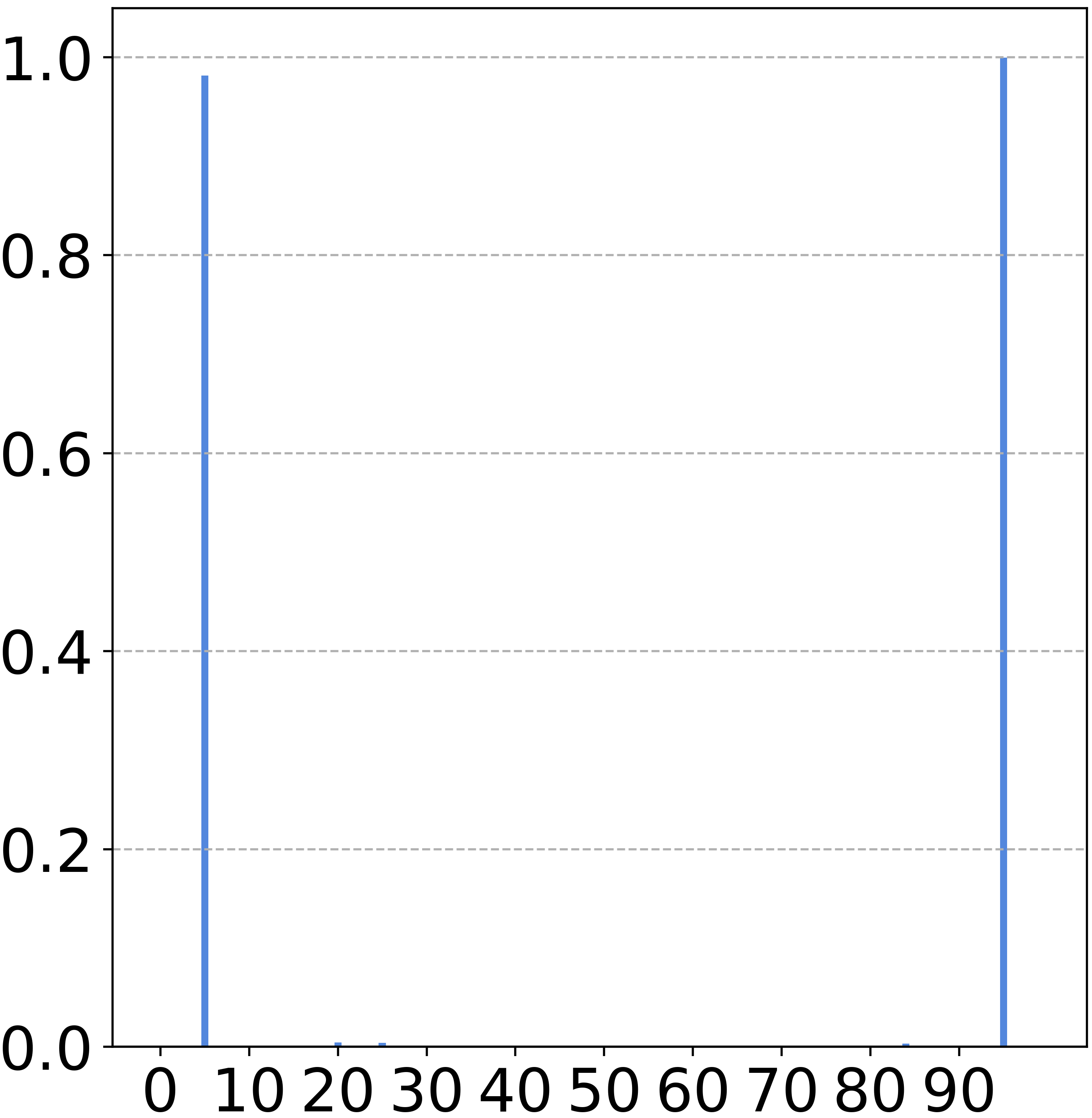}
		\caption{Class Probability}
		\label{fig:ClassProbability}
	\end{subfigure}
	\begin{subfigure}{0.157\textwidth}
		\includegraphics[width=\textwidth]{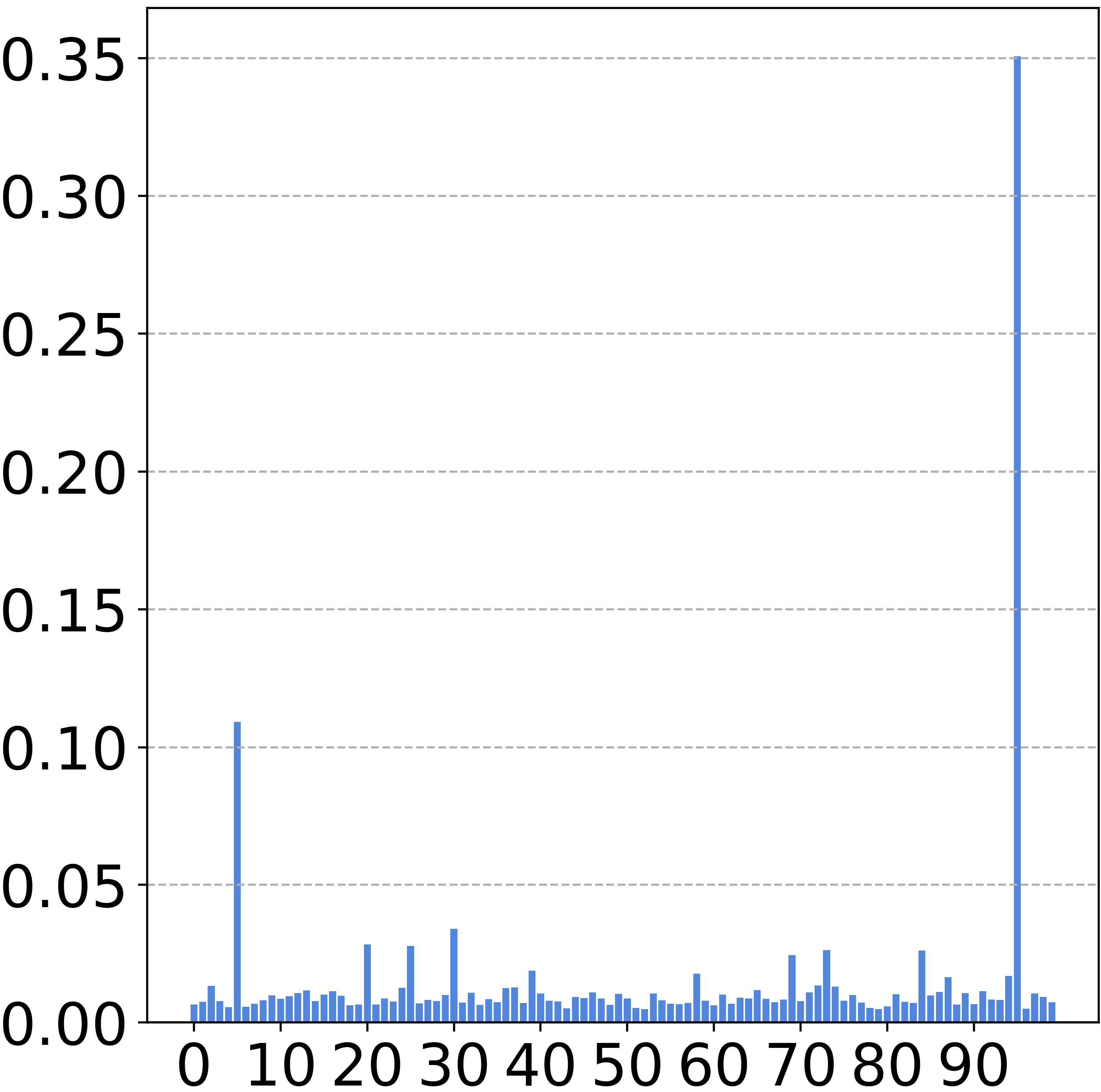}
		\caption{Softmax(Logits/$\tau$)}
		\label{fig:Softmax(Logits/T)}
	\end{subfigure}
	\caption{Differences between raw logits, class probabilities (softmaxed logits), and softmaxed logits with temperature ($\tau$) from the pre-trained teacher model (ResNet32x4). The sample was randomly selected from the CIFAR-100 \cite{cifar100} dataset.}
	\label{fig:logit_comparison}
\end{figure}
In knowledge distillation (KD), it is widely acknowledged that converting logits into probabilities using the \textbf{softmax} function with a temperature ($\tau$) enhances the amount of extracted information. Thus, let \(z_t^i\) denote the teacher’s logits and \(z_s^i\) the student’s logits, with \(C\) representing the number of classes:
\begin{gather}
	\mathcal{L}_{KD}^i = \tau^2 KL(z_s^i, z_t^i) \\
	KL(z_s^i, z_t^i) = \sum_{j=1}^{C} \sigma_j^t \log \frac{\sigma_j^t}{\sigma_j^s} \\
	\sigma_j^t = \text{softmax} \left( \frac{z_t^i}{\tau} \right)_j, \quad
	\sigma_j^s = \text{softmax} \left( \frac{z_s^i}{\tau} \right)_j
	\label{eq:kl}
\end{gather}
However, as illustrated in \cref{fig:logit_comparison}, after converting logits into probabilities, even with the temperature ($\tau$), the information in \cref{fig:Softmax(Logits/T)} is altered, and all values below zero are lost following the application of the \textbf{softmax} function. Thus, as shown in \cref{fig:Logits}, the logits contain both positive and negative values, whereas in \cref{fig:Softmax(Logits/T)}, the probabilities are strictly positive. While this transformation preserves the relative distribution of values, it significantly reduces information diversity. To enhance information diversity in logit distillation, we opted to use raw logits during the distillation process, discarding the KL divergence extension in favor of computing logit differences between teachers and students.\par
\begin{figure}[htbp]
	\centering
	\begin{subfigure}{0.18\textwidth}
		\includegraphics[width=\textwidth]{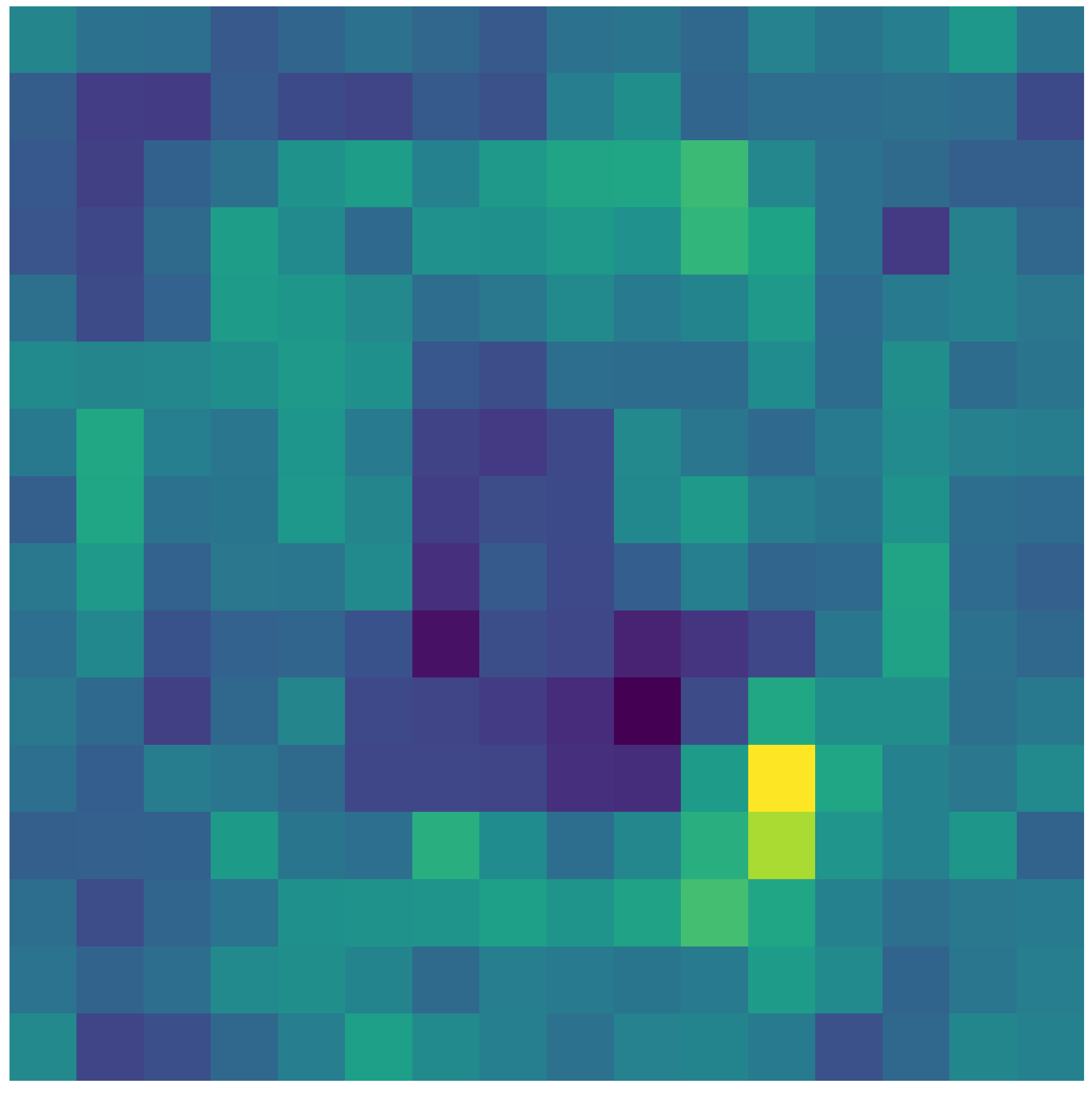}
		\caption{Feature map}
		\label{fig:Feature_map_vision}
	\end{subfigure}
	\begin{subfigure}{0.29\textwidth}
		\includegraphics[width=\textwidth]{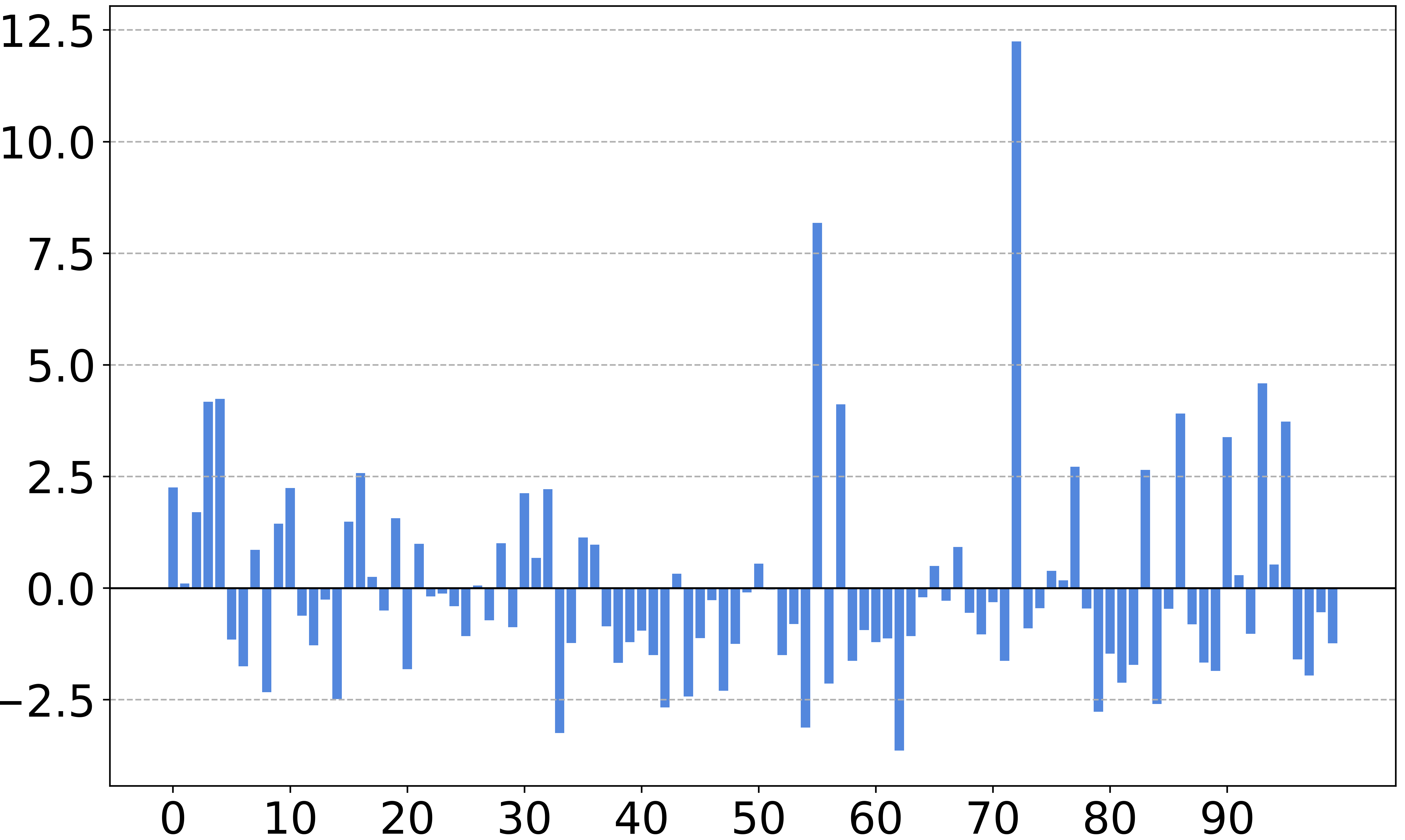}
		\caption{Logits}
		\label{Logits_vision}
	\end{subfigure}
	\caption{Visualization of the penultimate feature map (mean in channel-wise after activation) and raw logits.}
	\label{fig:feat_logits}
\end{figure}
\begin{figure*}[htbp]
	\centering
	\begin{subfigure}{0.49\textwidth}
		\includegraphics[width=\textwidth]{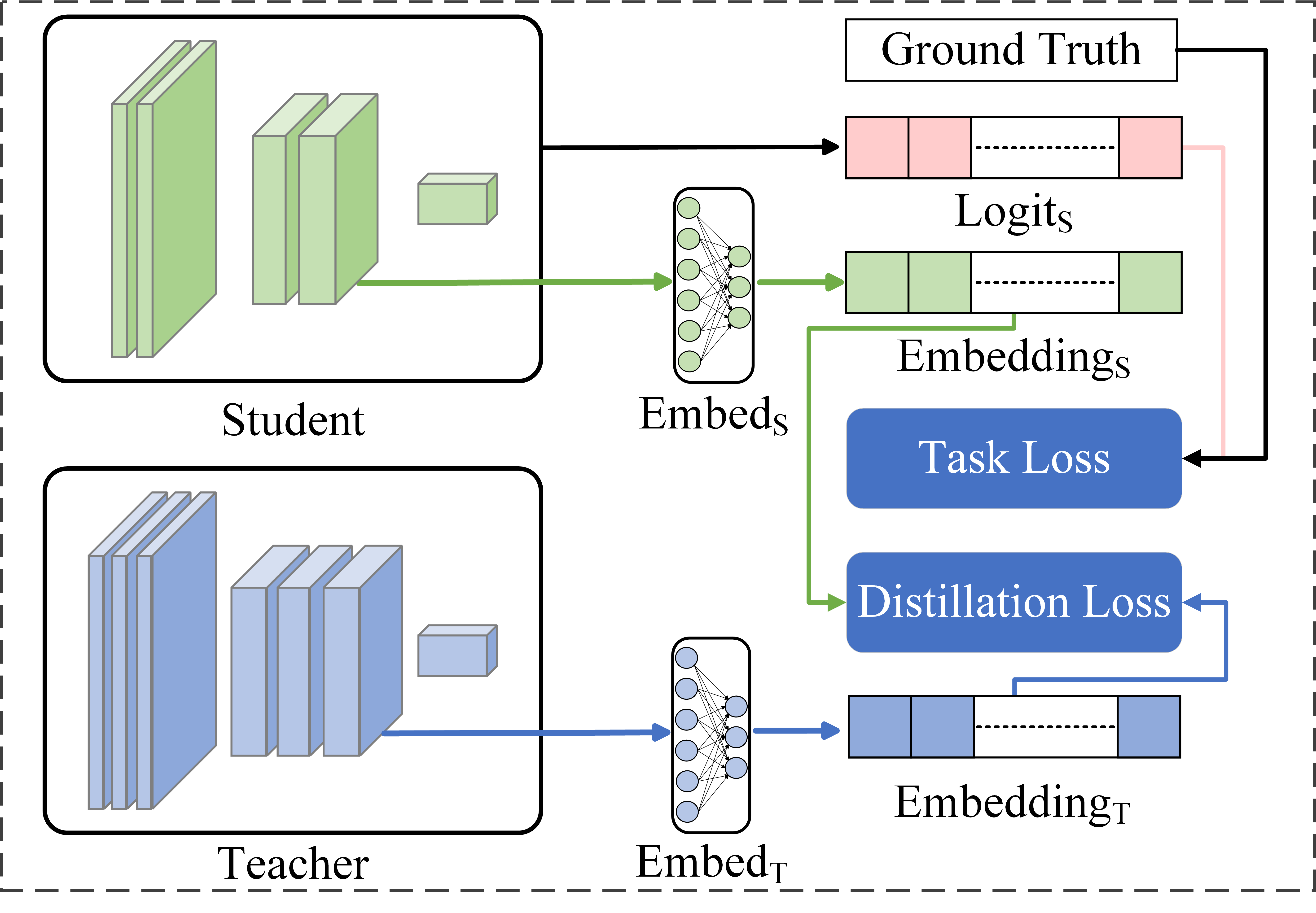}
		\caption{Feature distillation network architecture.}
		\label{fig:feature_distillation}
	\end{subfigure}
	\begin{subfigure}{0.49\textwidth}
		\includegraphics[width=\textwidth]{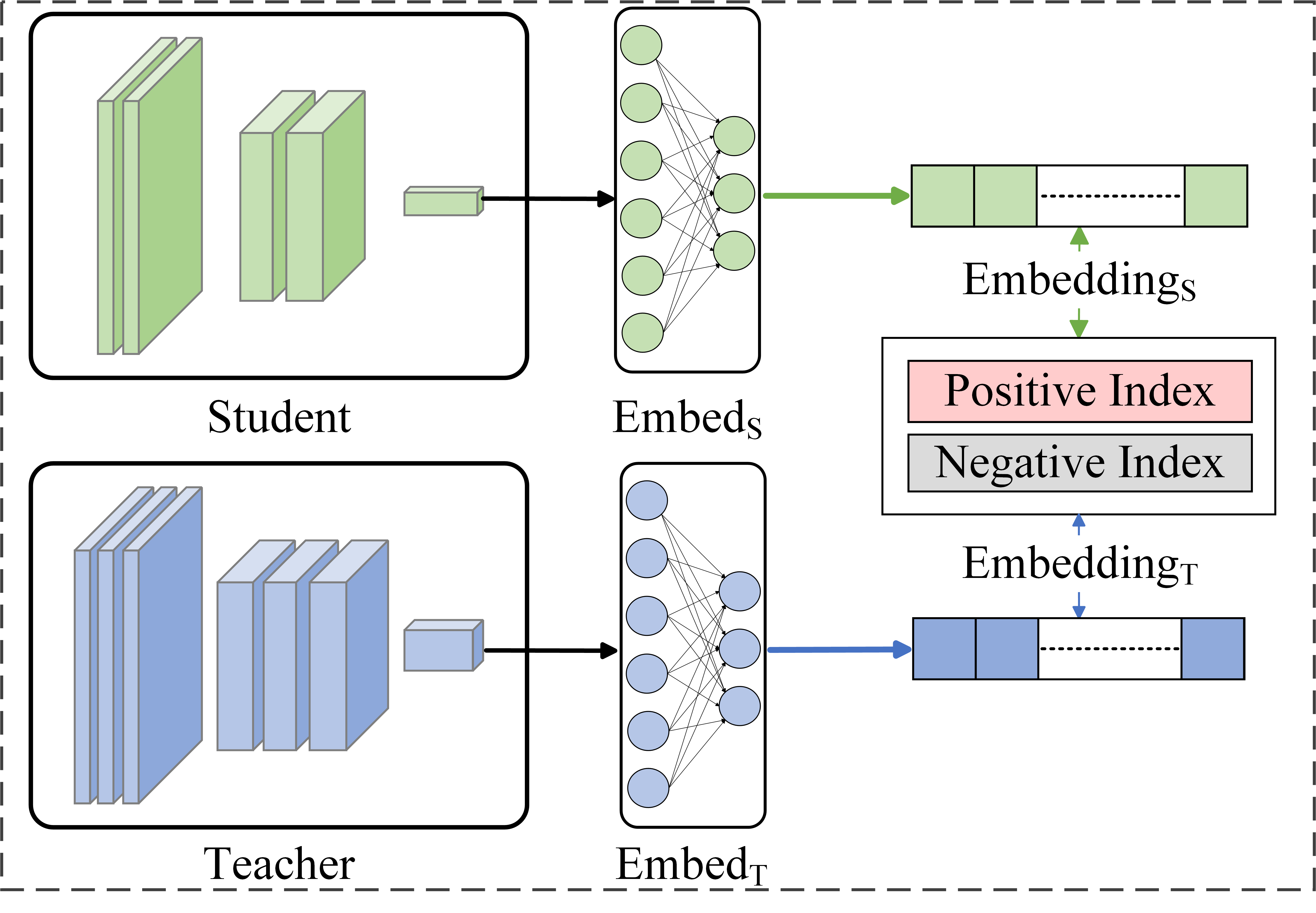}
		\caption{Feature distillation with contrastive learning, e.g., CRD\cite{crd}.}
		\label{fig:CRD}
	\end{subfigure}
	\begin{subfigure}{0.99\textwidth}
		\includegraphics[width=\textwidth]{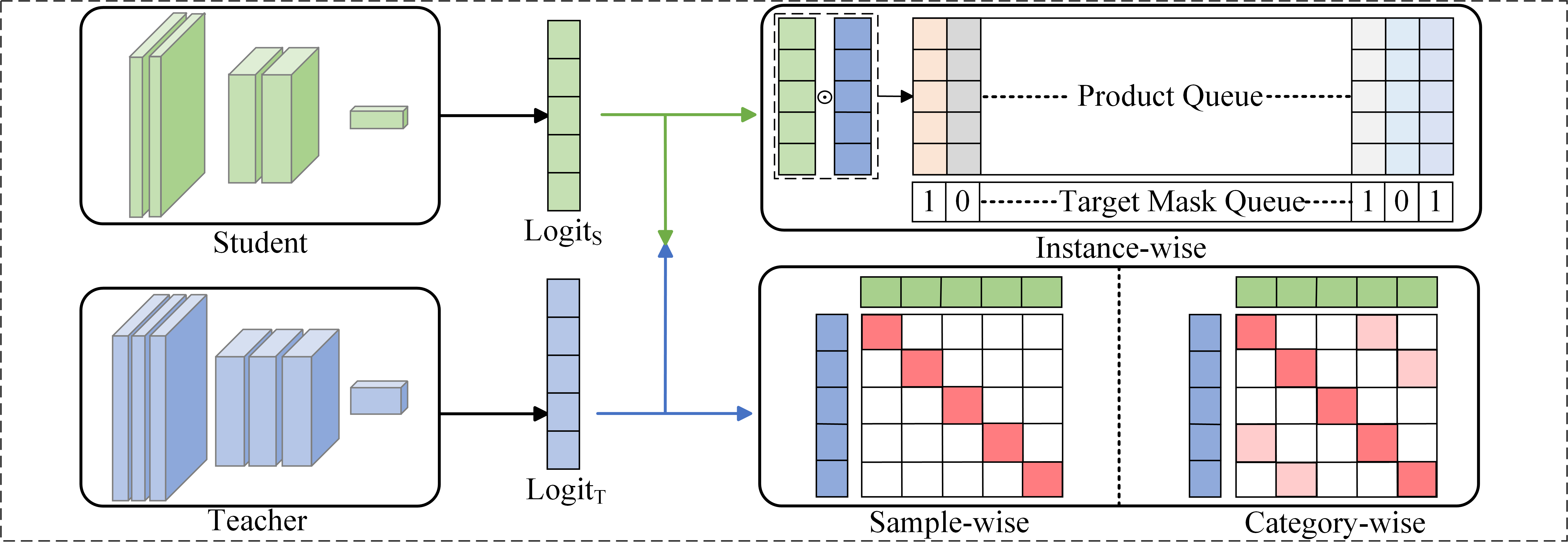}
		\caption{\textbf{MCLD} (No additional mapping modules, no additional positive or negative sample indexing, and multi-perspective utilization of logits).}
		\label{fig:MCLD}
	\end{subfigure}
\end{figure*}
Additionally, we visualize the penultimate layer of the pre-trained teacher model’s feature map and logits in \cref{fig:feat_logits}. It is well known that the feature map retains low-level detail information, whereas logits encode high-level semantic information. This semantic property of logits enables rapid classification of a given sample while also identifying classes that are similar or dissimilar to it. Moreover, through the study of \cite{moco,SimCLRv2}, we realized that knowledge distillation and contrastive learning have similar learning paradigms. Thus, to leverage the semantic properties of logits while preserving their information diversity by avoiding conversion to probabilities via the softmax function, we adopted contrastive learning for logit distillation. Furthermore, although in feature distillation, the influential CRD \cite{crd} also applied contrastive learning to knowledge distillation and achieved state-of-the-art performance at the time. However, MCLD is quite different from traditional logit distillation methods, as elaborated in \cref{sec:methodology}.
\section{Methodology}
\label{sec:methodology}
In this section, we first provide a brief review of the feature distillation (especially methods that combine with contrastive learning) to facilitate a better understanding of MCLD. In \cref{sec:multi-perspective}, we introduce the various modules proposed in MCLD. Finally, we describe the complete training process in \cref{sec:loss-function}.
\subsection{Review of Feature Distillation}
\label{sec:reviewcrd}
Most feature distillation methods \cite{similarity,crd,review,reuse,class} conjectured that the teacher’s feature map more effectively transfers more information rather than solely conveying knowledge about output class probabilities, as shown in \cref{fig:feat_logits}. Some feature distillation methods combine with contrastive learning, as \cite{crd,wocrd} selects the feature map at the last or penultimate layer as the contrastive objective to maximize the transfer of information during the distillation process. Additionally, since feature distillation methods use the feature map as a transfer representation, they require an additional mapping module ($\textbf{Embed}_\textbf{S}$, $\textbf{Embed}_\textbf{T}$ in \cref{fig:feature_distillation}) to align the feature map dimensions of the teacher and student. In addition, for example, \cite{crd}, uses contrastive learning, which requires to additionally sample the indexes of positive and negative samples ($\textbf{Positive Index}$, $\textbf{Negative Index}$ in \cref{fig:CRD}) in the data sampling process to distinguish the different samples, which increases the time of data sampling.
\subsection{Multi-perspective Distillation}
\label{sec:multi-perspective}
Compared to feature distillation methods \cite{review,reuse,class,FCFD}, MCLD directly utilizes logits for distillation without requiring additional mapping modules ($\text{Embed}_\text{S}$, $\text{Embed}_\text{T}$ in most feature distillation methods). Additionally, MCLD directly utilizes ground truth labels to generate the \textbf{Target Mask} in \cref{fig:MCLD}, distinguishing positive and negative samples, and does not require additional sample indexes (Positive Index, Negative Index), which reduces data sampling time. Furthermore, to effectively leverage the semantic properties of logits and improve information diversity in logit distillation, MCLD utilizes logits from three perspectives: \textbf{Instance-wise}, \textbf{Sample-wise}, and \textbf{Category-wise} in \cref{fig:MCLD}.
\subsubsection{Instance-wise CLD}
\label{sec:Instance-wise}
To capture both similarities and differences between samples, we consider each sample in the dataset as a distinct instance. Our objective is to bring \(z_s^i\) and \(z_t^i\) closer together while pushing \(z_s^i\) and \(z_t^j\) further apart. Using InfoNCE \cite{infonce} as the loss function, its formula is as follows:
\begin{gather}
	\sigma_i = z_s^i \cdot z_t^i\quad \sigma_j = z_s^i \cdot z_t^j \\ 
	\mathcal{L}_{InfoNCE} = -\log\frac{\exp\left(\sigma_i / \tau\right)}{\sum_{j=0}^{K} \exp\left(\sigma_j / \tau\right)}
	\label{eq:infonce_instance}
\end{gather}
where \{\(z_s^i\), \(z_t^i\), \(z_t^j\)\} $\in \mathbb{R}^{B \times C}$, \{\(\sigma_i\), \(\sigma_j\)\} $\in \mathbb{R}^{B}$, \(B\) represents the batch size. Additionally, we employ a queue ($\textbf{Product Queue}$ in \cref{fig:MCLD}) to store all \(\sigma_j\) values in the training dataset. The queue is defined as \(\mathbb{R}^{C \times K} \), where \( K \) denotes the queue length, which can be set independently as a hyperparameter. This structure enables each training epoch to generate \( K \) negative logits pairs (\(\sigma_j\)) via the queue, along with one positive logit pair (\(\sigma_i\)). The queue structure enables continuous updates to the stored \(\sigma_j\) values with each mini-batch.
\begin{gather}
	\boldsymbol{\gamma} =
	\begin{cases}
		1, & \text{if } y_s^i \neq y_t^j \\
		0, & \text{if } y_s^i = y_t^j
	\end{cases}
	\label{eq:target_mask}
\end{gather}
Meanwhile, \(y_s^i\) and \(y_t^j\) represent the ground truth labels. We employ another queue ($\textbf{Target Mask Queue}$ in \cref{fig:MCLD}) to store the target masks \(\boldsymbol{\gamma}\), ensuring that samples from the same category as \(z_t^j\) are treated as positive samples instead of negative samples for \(z_s^i\). This queue is updated synchronously with the $\textbf{Product Queue}$.\par
Next, we formulate \cref{eq:infonce_instance} as a \((K+1)\)-way classification task. We define a one-hot label \(y_{inst}\), where the positive logits pair is assigned a value of 1 at the 0-th position, while the \(K\) negative logits pairs are assigned a value of 0 from the 1st to \(K\)th position. Thus, \cref{eq:infonce_instance} can be reformulated as a Cross-Entropy Loss (CE) function, as follows:
\begin{gather}
	\mathcal{L}_{Inst} = -\sum_{i} y_{inst} \log(\frac{\exp\left(\sigma_i/\tau\right)}{\sum_j \exp\left(\boldsymbol{\gamma} * \sigma_j/\tau\right)})
	\label{eq:Inst}
\end{gather}
Instance-wise CLD enables a simple and efficient comparison between the logits of the student and teacher across all samples in the training dataset. In contrast to most logit distillation methods, which typically rely on mini-batch comparisons (e.g., batch sizes of 64, 128, 256), our method performs a 1-to-\(N\) comparison, where \(N\) represents all training samples, excluding those from the same category as the current sample. This approach enhances logit distillation performance by comparing a large number of samples.
\subsubsection{Sample-wise CLD}
\label{sec:Sample-wise}
In \cref{sec:Instance-wise}, Instance-wise CLD can help students discern the differences between samples of different categories in the training dataset. To further utilize the potential of logits, we propose Sample-wise CLD to promote similarity between the same samples. The similarity between \(z_s^i\) and \(z_t^i\) is computed as follows:
\begin{gather}
	\eta_i = z_s^i \cdot {z_t^i}^T
	\label{eq:similarity}
\end{gather}
where $\eta_i \in \mathbb{R}^{B \times B}$ is the similarity matrix. The diagonal elements of \(\eta_i\) represent the similarity between \(z_s^i\) and \(z_t^i\) for the same sample in each batch. Thus, we treat the same sample comparison as a \(B\)-way classification task, so we define a label \(y_{samp}\) consisting of natural numbers ranging from 0 to \(B-1\). The loss function is computed as follows: 
\begin{gather}
	\mathcal{L}_{Samp} = \mathcal{L}_{ce}(\eta_i /\tau, y_{samp})
	\label{eq:similarity_loss}
\end{gather}
\subsubsection{Category-wise CLD}
\label{sec:Category-wise}
Instance-wise CLD helps students discern the differences between samples across different categories, while Sample-wise CLD captures the similarity between identical samples from the teacher and student models. Additionally, different samples within the same category should be treated as positive samples to learn both similarities and differences, as discussed in \cite{sup}.
Therefore, we introduce Category-wise CLD as a complement to \cref{eq:similarity_loss}. To effectively determine whether samples belong to the same category, we utilize the ground-truth labels from the dataset. Within each mini-batch, samples from the same category are designated as positive samples, while those from different categories are treated as negative samples. Accordingly, we define Category-wise CLD as follows:
\begin{gather}
	\psi_p = z_s^i \cdot z_t^p\quad \psi_n = z_s^i \cdot z_t^n \\
	\mathcal{L}_{Cate} = -{\frac{1}{P}}\sum\limits_{p\in P}\log\frac{\exp(\psi_p/\tau)}{\sum\limits_{n\in N}\exp(\psi_n/\tau)}
	\label{eq:category}
\end{gather}
where $\psi_p$ represents samples from the same category as $z_s^i$ but not the same sample, and $\psi_n$ denotes samples from different categories. Here, $P$ and $N$ denote the number of positive and negative samples in each batch, respectively.
\begin{table*}[htbp]
	\centering
	\caption{The Top-1 accuracy (\%) of knowledge distillation methods on the CIFAR-100 \cite{cifar100} validation set. The teacher and student models have a \textbf{heterogeneous structure}. The methods are categorized based on the distillation approach: feature distillation and logit distillation. The best and second-best results across all methods are highlighted in \textbf{bold} and \underline{underlined}, respectively. The * symbol denotes our reproduced results under the same experimental setting, while / indicates an inability to reproduce results for the current model pair. The $\blacktriangledown$ symbol signifies that LSKD+MLKD \cite{lskd} completely failed to converge within 240 epochs. All results are reported as the average over three trials.}
	\label{tab:cifar100-dif-table}
	\resizebox{\textwidth}{!}{%
		\begin{tabular}{@{}ccccccccc@{}}
			\toprule
			& \multirow{2}{*}{Teacher} & ResNet32×4    & ResNet32×4 & ResNet32×4 & WRN-40-2  & WRN-40-2     & VGG13        & ResNet50     \\
			Distillation              &            & 79.42 & 79.42 & 79.42 & 75.61 & 75.61 & 74.64 & 79.34 \\
			Manner & \multirow{2}{*}{Student} & SHN-V2 & WRN-16-2   & WRN-40-2   & ResNet8×4 & MN-V2 & MN-V2 & MN-V2 \\
			&            & 71.82 & 73.26 & 75.61 & 72.50 & 64.60 & 64.60 & 64.60 \\ \midrule
			& CRD \cite{crd}       & 75.65 & 75.65 & 78.15 & 75.24 & 70.28 & 69.73 & 69.11 \\
			& ReviewKD \cite{review}   & 77.78 & 76.11 & 78.96 & 74.34 & \underline{71.28} & 70.37 & 69.89 \\
			\multirow{2}{*}{Feature}
			& SimKD \cite{reuse}     & 78.39 & \underline{77.17} & \underline{79.29} & 75.29 & 70.10 & 69.44 & 69.97 \\
			& CAT-KD \cite{class}    & \underline{78.41} & 76.97 & 78.59 & 75.38 & 70.24 & 69.13 & \underline{71.36} \\
			& FCFD* \cite{FCFD}  & 78.25 & 76.73 & 79.27 & 76.26 & / & \underline{70.69} & 70.89 \\
			\midrule
			& KD \cite{kd}        & 74.45 & 74.90 & 77.70 & 73.07 & 68.36 & 67.37 & 67.35 \\
			& CTKD \cite{ctkd}      & 75.37 & 74.57 & 77.66 & 74.61 & 68.34 & 68.50 & 68.67 \\
			& DKD \cite{dkd}       & 77.07 & 75.70 & 78.46 & 75.56 & 69.28 & 69.71 & 70.35 \\
			\multirow{2}{*}{Logit}   
			& DOT+DKD* \cite{dot}  & 77.41 & 75.74 & 78.33 & 75.61  & 68.99  & 61.43 & 70.26 \\
			& LSKD+MLKD* \cite{lskd}  & 72.13 & 74.06 & 78.38 & 61.52 & 68.11 & 60.58 & $19.63^{\blacktriangledown}$ \\
			\multicolumn{1}{l}{}      & LSKD+DKD* \cite{lskd} & 77.09 & 75.75 & 78.60 & 75.93  & 69.28 & 69.98 & 70.78 \\
			& WTTM* \cite{TTM}    & 76.59 & 76.04 & 78.45 & \underline{76.40} & 68.81 & 69.18 & 69.95 \\
			& \textbf{Ours}       & \textbf{78.87} & \textbf{77.47} & \textbf{80.37} & \textbf{77.16} & \textbf{71.91} & \textbf{70.96} & \textbf{72.83} \\ \bottomrule
		\end{tabular}%
	}
\end{table*}
\subsection{Loss Function}
\label{sec:loss-function}
Thus far, we have analyzed logits from multiple perspectives, extracting maximal information from the teacher model’s logits. We argue that distinguishing between different samples within the same category requires the model to possess a certain level of discriminative ability. Therefore, we design a warm-up learning strategy to enable \(\mathcal{L}_{Cate}\) to play a more significant role in the later stages of training. The loss function of MCLD is defined as follows:
\begin{gather}
	\mathcal{L}_{MCLD} = \mathcal{L}_{Inst} + \mathcal{L}_{Samp} + \omega * \mathcal{L}_{Cate}
	\label{eq:mpcld_loss}
\end{gather}
Notably, MCLD does not require any loss weight parameters or deliberate tuning of hyperparameters during training (\(\omega\) follows the same warm-up learning strategy as in \cite{dkd, dot}. As training progresses, \(\omega\) increases from 0 to 1 by the end of the warm-up epoch). This further underscores the effectiveness and superiority of MCLD, demonstrating that it is a simple, general, and efficient logit distillation method. We empirically examine and validate the effects of the three MCLD modules and the warm-up learning strategy in \cref{sec:ablation}.
\section{Experiments}
\label{sec:experiments}
\textbf{Dataset.} Our experiments primarily focus on image classification. CIFAR-100 \cite{cifar100} is a widely used image classification dataset, consisting of 60,000 color images, each measuring 32×32 pixels. It contains 50,000 training images and 10,000 test images, distributed across 100 classes. ImageNet \cite{imagenet} is a large-scale classification dataset containing 1.28 million training images and 50,000 validation images, distributed across 1,000 categories.\par
\noindent\textbf{Baselines.} Over the past decades, most state-of-the-art methods have relied on feature distillation. We begin by comparing our method with these methods, including CRD \cite{crd}, ReviewKD \cite{review}, SimKD \cite{reuse}, CAT-KD \cite{class}, WTTM \cite{TTM} and FCFD \cite{FCFD}. Since MCLD is a logit distillation method, we also conduct a more comprehensive comparison with classic KD \cite{kd} and recent state-of-the-art logit distillation methods, including CTKD \cite{ctkd}, DKD \cite{dkd}, DOT \cite{dot}, MLKD \cite{mlkd}, and LSKD \cite{lskd}.\par
\noindent\textbf{Training details.} We followed the same experimental settings as previous work \cite{crd,moco,dkd}. For all experiments, we used the SGD optimizer with a batch size of 64, training for 240 epochs on CIFAR-100 and 100 epochs on ImageNet. More details are attached in the supplements.\par
\begin{table*}[htbp]
	\centering
	\caption{Top-1 accuracy (\%) of knowledge distillation methods on the CIFAR-100 \cite{cifar100} validation set. The teacher and student models have a \textbf{homogeneous structure}. The KD methods are categorized by distillation approach: feature distillation and logit distillation. The best and second-best results across all methods are highlighted in \textbf{bold} and \underline{underlined}, respectively. Methods marked with * indicate reproduced results under the same hardware and experimental conditions.}
	\label{tab:cifar100-same}
	\resizebox{\textwidth}{!}{%
		\begin{tabular}{@{}ccccccccc@{}}
			\toprule
			& \multirow{2}{*}{Teacher} & ResNet32×4 & VGG13 & WRN-40-2 & WRN-40-2 & ResNet56 & ResNet110 & ResNet110 \\
			Distillation              &            & 79.42 & 74.64 & 75.61 & 75.61 & 72.34 & 74.31 & 74.31 \\
			Manner & \multirow{2}{*}{Student} & ResNet8x4  & VGG8  & WRN-40-1 & WRN-16-2 & ResNet20 & ResNet32  & ResNet20  \\
			&            & 72.50 & 70.36 & 71.98 & 73.26 & 69.06 & 71.14 & 69.06 \\ \midrule
			& CRD \cite{crd}       & 75.51 & 73.94 & 74.14 & 75.48 & 71.16 & 73.48 & 71.46 \\
			& ReviewKD \cite{review}  & 75.63 & 74.84 & 75.09 & 76.12 & 71.89 & 73.89 & 71.34 \\
			\multirow{2}{*}{Feature}
			& SimKD \cite{reuse}     & \underline{78.08} & 74.80 & 74.53 & 75.53 & 71.05 & 73.92 & 71.06 \\
			& CAT-KD \cite{class}    & 76.91 & 74.65 & 74.82 & 75.60 & 71.62 & 73.62 & 71.37 \\  
			& FCFD* \cite{FCFD}        & 76.98 & \underline{74.96} & \underline{75.53} & \underline{76.34} & 71.68 & 73.75 & \underline{71.92} \\
			\midrule
			& KD \cite{kd}        & 73.33 & 72.98 & 73.54 & 74.92 & 70.66 & 73.08 & 70.67 \\
			& CTKD+KD \cite{ctkd}   & 76.67 & 73.52 & 73.93 & 75.45 & 71.19 & 73.52 & 70.99 \\
			& DKD \cite{dkd}       & 76.32 & 74.68 & 74.81 & 76.24 & 71.97 & 74.11 & 71.06 \\
			\multirow{2}{*}{Logits}   
			& DOT+DKD* \cite{dot} & 76.12 & 73.93 & 74.13 & 75.36 & \underline{72.11}      & 71.39       & 69.43      \\
			& LSKD+MLKD* \cite{lskd}  & 71.21 & 57.88 & 72.45 & 74.55 & 69.83 & 54.55 & 66.31 \\
			\multicolumn{1}{l}{}      & LSKD+DKD* \cite{lskd}  & 76.79 & 74.80 & 74.73 & \underline{76.34} & \textbf{72.31} & 73.89 & 71.39 \\
			& WTTM* \cite{TTM}    & 76.24 & 74.27 & 74.08 & 76.29 & 71.65 & \underline{74.26} & 71.16 \\
			& \textbf{Ours}       & \textbf{78.19} & \textbf{75.36} & \textbf{76.18} & \textbf{76.37} & 71.75 & \textbf{74.27} & \textbf{72.28} \\ \bottomrule
		\end{tabular}%
	}
\end{table*}
\subsection{Main Results}
\textbf{CIFAR-100 image classification.} As in previous works \cite{crd,review,dkd,dot}, we use the CIFAR-100 dataset as the primary benchmark to compare the performance of various KD methods under different network settings, as presented in \cref{tab:cifar100-dif-table} and \cref{tab:cifar100-same}. \cref{tab:cifar100-dif-table} presents the results of experiments where the teacher and student models have a heterogeneous structure, while \cref{tab:cifar100-same} shows results for a homogeneous structure.\par
Remarkably, MCLD consistently improves performance across most teacher-student pairs, achieving a notable improvement of \textcolor[HTML]{EF3340}{1$\sim$2\%} compared to state-of-the-art logit distillation methods. This strongly demonstrates the effectiveness and superiority of our MCLD. Furthermore, MCLD still achieves comparable or even superior performance compared to state-of-the-art feature distillation methods. To ensure fairness and impartiality in our experiments, we reproduce all methods using 240 epochs, following the setting in \cite{crd,review,dkd,dot}. However, LSKD+MLKD \cite{lskd}, when trained for 240 epochs (480 epochs in the original report), exhibited significant performance degradation in \cref{tab:cifar100-dif-table} and \cref{tab:cifar100-same}, and failed to converge in some cases. In addition, we also reproduce the same epoch (480 epochs) as in the original report. More results are attached in the supplements.\par
\noindent\textbf{ImageNet image classification.} Top-1 and Top-5 accuracy are evaluated and compared in \cref{tab:imagenet}. In the first case, ResNet34/ResNet18, there was minimal difference between the teacher and student models. In the second case, ResNet50/MobileNetV1, there was a greater disparity between the teacher and student models, with the teacher achieving better performance. On the large-scale dataset ImageNet, MCLD consistently outperforms most state-of-the-art distillation methods.
\begin{table}[htbp]
	\centering
	\caption{Top-1 and Top-5 accuracy (\%) on the ImageNet \cite{imagenet} validation set. The best and second-best results are emphasized in \textbf{bold} and \underline{underlined}.}
	\label{tab:imagenet}
	\resizebox{\columnwidth}{!}{%
		\begin{tabular}{@{}ccccc@{}}
			\toprule
			Teacher/Student & \multicolumn{2}{c}{ResNet34/ResNet18} & \multicolumn{2}{c}{ResNet50/MN-V1} \\
			Acc      & Top-1 & Top-5 & Top-1 & Top-5 \\ \midrule
			Teacher  & 73.31 & 91.42 & 76.16 & 92.86 \\
			Student  & 69.75 & 89.07 & 68.87 & 88.76 \\ \midrule
			CRD \cite{crd}    & 71.17 & 90.13 & 71.37 & 90.41 \\
			ReviewKD \cite{review} & 71.61 & 90.51 & 72.56 & 91.00 \\
			SimKD \cite{reuse}   & 71.59 & 90.48 & 72.25 & 90.86 \\
			CAT-KD \cite{class}  & 71.26 & 90.45 & 72.24 & 91.13 \\
			FCFD \cite{FCFD} & \textbf{72.24} & \underline{90.74} & \underline{73.37} & \underline{91.35} \\ \midrule
			KD \cite{kd}      & 71.03 & 90.05 & 70.50 & 89.80 \\
			CTKD+KD \cite{ctkd,kd} & 71.38 & 90.27 & 71.16 & 90.11 \\
			DKD \cite{dkd}     & 71.70 & 90.41 & 72.05 & 91.05 \\
			LSKD+DKD \cite{lskd,dkd} & 71.88 & 90.58 & 72.85 & 91.23 \\
			WTTM \cite{TTM}      & \underline{72.19} & n/a & 73.09 & n/a \\
			\textbf{Ours}     & 71.90 & \textbf{90.91} & \textbf{73.63} & \textbf{91.98}      \\
			\bottomrule
		\end{tabular}%
	}
\end{table}
\subsection{Ablation Studies}
\label{sec:ablation}
To provide a comprehensive understanding of MCLD, we conducted extensive ablation studies on the three modules presented in this paper. The results are shown in \cref{tab:diff-modules}. Additionally, to intuitively observe the impact of the warm-up learning strategy on MCLD, refer to the ablation results in \cref{tab:warm-up} for more detailed insights.\par
As shown in \cref{tab:diff-modules}, all three modules proposed in this paper improve the baseline, and combining any two modules leads to better performance, indicating that one can compensate for the deficiencies of another, thereby contributing to overall improvement.
\begin{table}[htbp]
	\centering
	\caption{Top-1 accuracy (\%) on the CIFAR-100 validation set. We set ResNet8x4 as the student and ResNet32x4 as the teacher. The $\Delta$ symbol represents the performance improvement over the baseline.}
	\label{tab:diff-modules}
	\resizebox{\columnwidth}{!}{
		\begin{tabular}{@{}cccc|cl@{}}
			\toprule
			Instance-wise & Sample-wise & Category-wise & Acc & $\Delta$ \\ \midrule
			$\times$ & $\times$ & $\times$  & 72.50  &  - \\
			$\checkmark$ & $\times$ & $\times$  & 77.68 & \textcolor[HTML]{EF3340}{+5.18}  \\
			$\times$ & $\checkmark$ & $\times$  & 77.76 & \textcolor[HTML]{EF3340}{+5.26}  \\
			$\times$ & $\times$ & $\checkmark$  & 77.09 & \textcolor[HTML]{EF3340}{+4.59}  \\ \midrule
			$\checkmark$ & $\checkmark$ & $\times$ & 77.83 & \textcolor[HTML]{EF3340}{+5.33} \\
			$\checkmark$ & $\times$ & $\checkmark$ & 78.05 & \textcolor[HTML]{EF3340}{+5.55} \\
			$\times$ & $\checkmark$ & $\checkmark$ & 77.71 & \textcolor[HTML]{EF3340}{+5.21} \\ \midrule
			$\checkmark$ & $\checkmark$ & $\checkmark$ & 78.19 & \textcolor[HTML]{EF3340}{+5.69} \\ \bottomrule
		\end{tabular}
	}
\end{table}
\begin{table}[htbp]
	\centering
	\caption{Top-1 accuracy (\%) on the CIFAR-100 \cite{cifar100} validation set with different $\omega$ End Epoch.}
	\label{tab:warm-up}
	\resizebox{\columnwidth}{!}{%
		\begin{tabular}{@{}c|cccccl@{}}
			\toprule
			$\omega$ End Epoch & 1 & 155 & 185 & 215 & 240 \\ \midrule
			Top-1 Acc & 77.72  & \textbf{78.19} & 77.89 & 77.74 & 77.90 \\ \midrule
			$\omega$ End Epoch & 50 & 100 & 150 & 180 & 210 \\ \midrule
			Top-1 Acc & 77.87  & 77.98 & \underline{78.05} & 78.01 & 77.94 \\ \bottomrule
		\end{tabular}%
	}
\end{table}
In addition, as shown in \cref{tab:warm-up}, fixing the loss weight of the Category-wise CLD module to 1 (with $\omega$ reaching 1 at the final epoch) at the beginning of training is less effective than gradually increasing its weight throughout training. This occurs because, in the early stages of training, the model’s discriminative ability is weak, making it difficult to distinguish differences between samples of the same category. Once the model acquires a certain level of discriminative ability later in training, the effectiveness of the Category-wise CLD module can be fully realized. The values 150, 180, and 210 correspond to the settings in \cite{crd,dkd,dot,FCFD}, where the learning rate is adjusted at these epochs. To ensure consistency, we set the epochs to 155, 185, and 215, aligning with the average training accuracy of the student model. More results of ablation studies are attached in the supplements.
\begin{figure*}[htbp]
	\centering
	\begin{subfigure}{0.185\textwidth}
		\includegraphics[width=\textwidth]{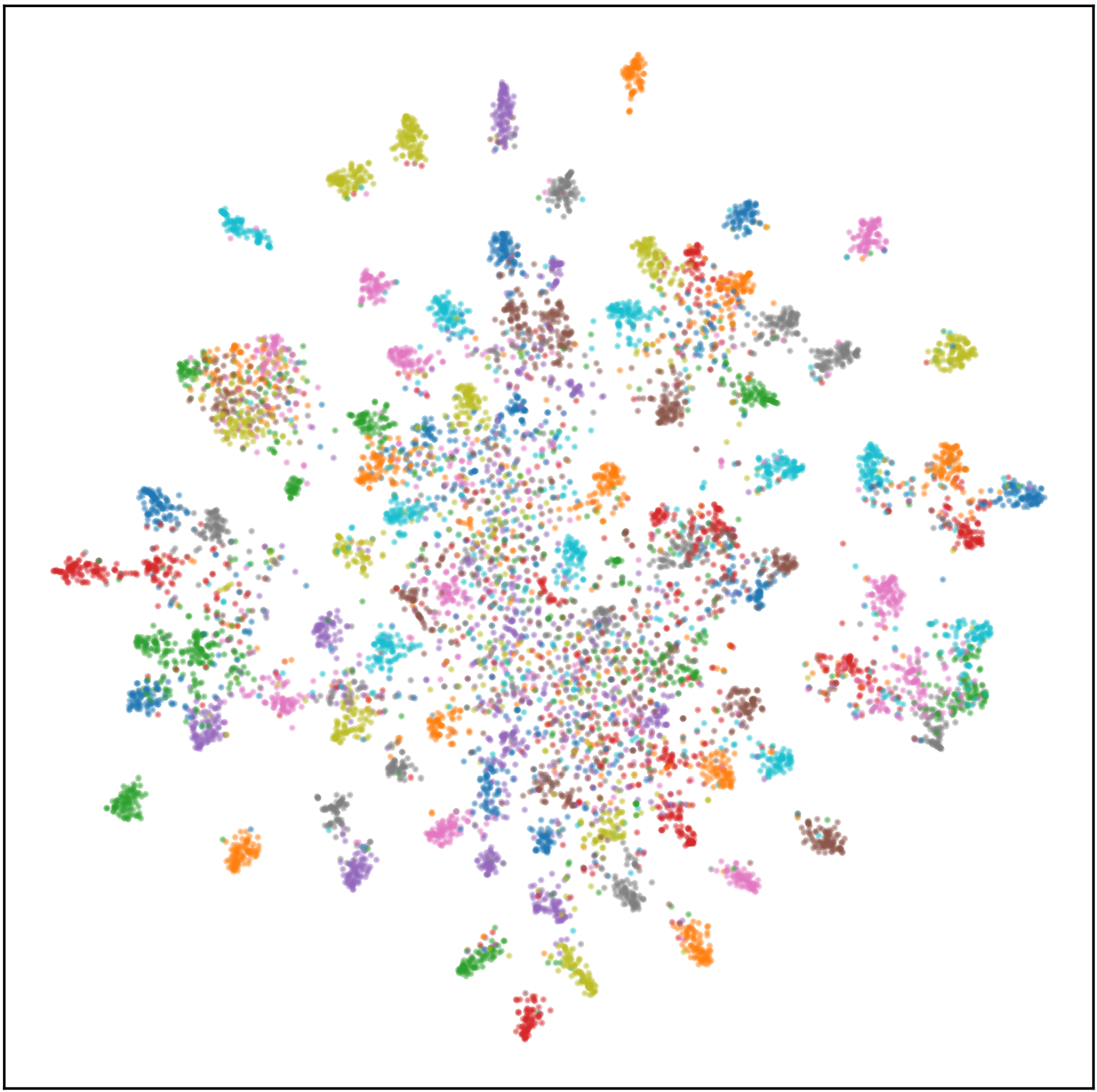}
		\caption{DOT+DKD \cite{dot}}
		\label{fig:dot+dkd_resnet8x4_tsne}
	\end{subfigure}
	\begin{subfigure}{0.185\textwidth}
		\includegraphics[width=\textwidth]{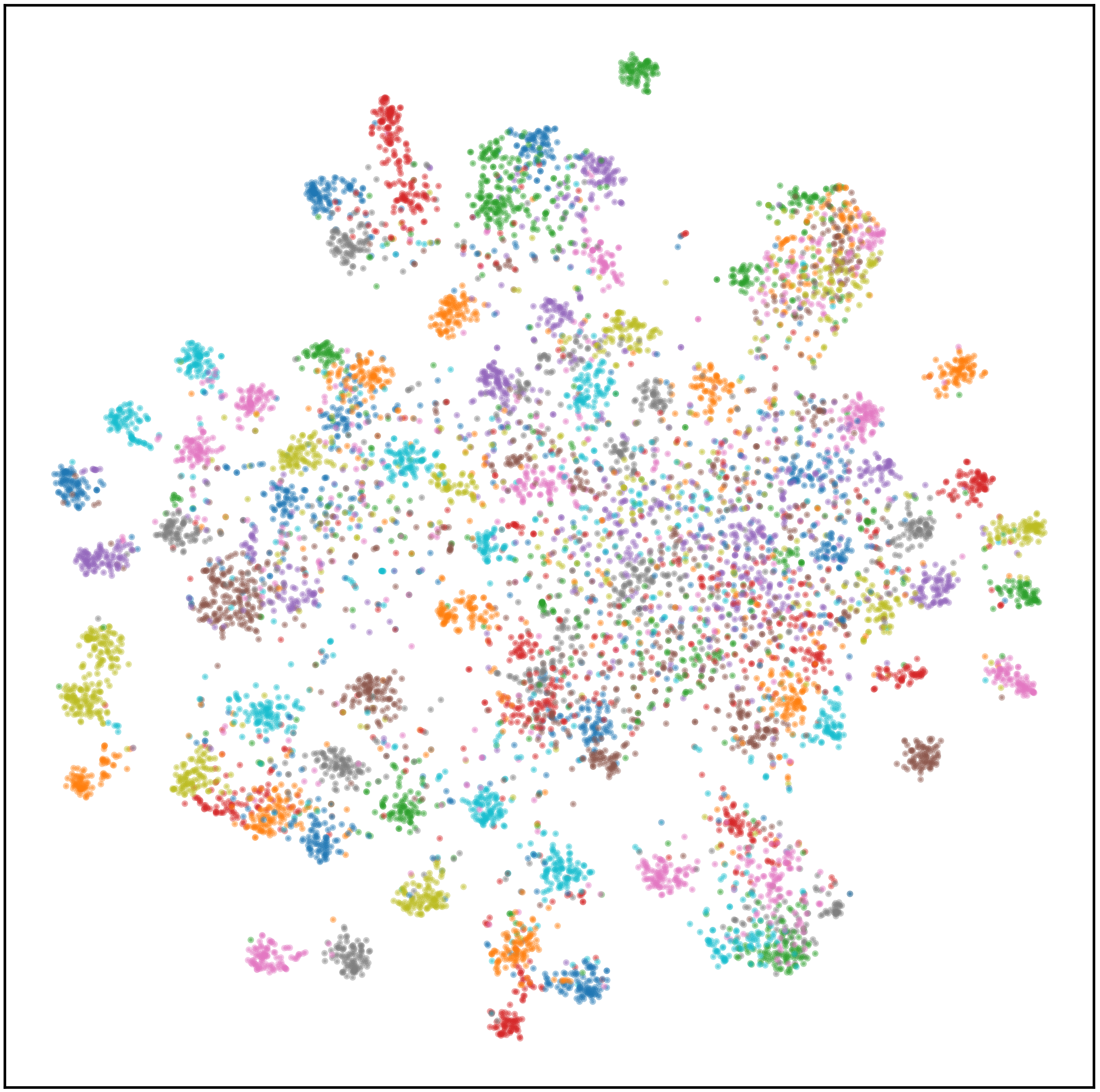}
		\caption{FCFD \cite{FCFD}}
		\label{fig:fcfd_resnet8x4_tsne}
	\end{subfigure}
	\begin{subfigure}{0.185\textwidth}
		\includegraphics[width=\textwidth]{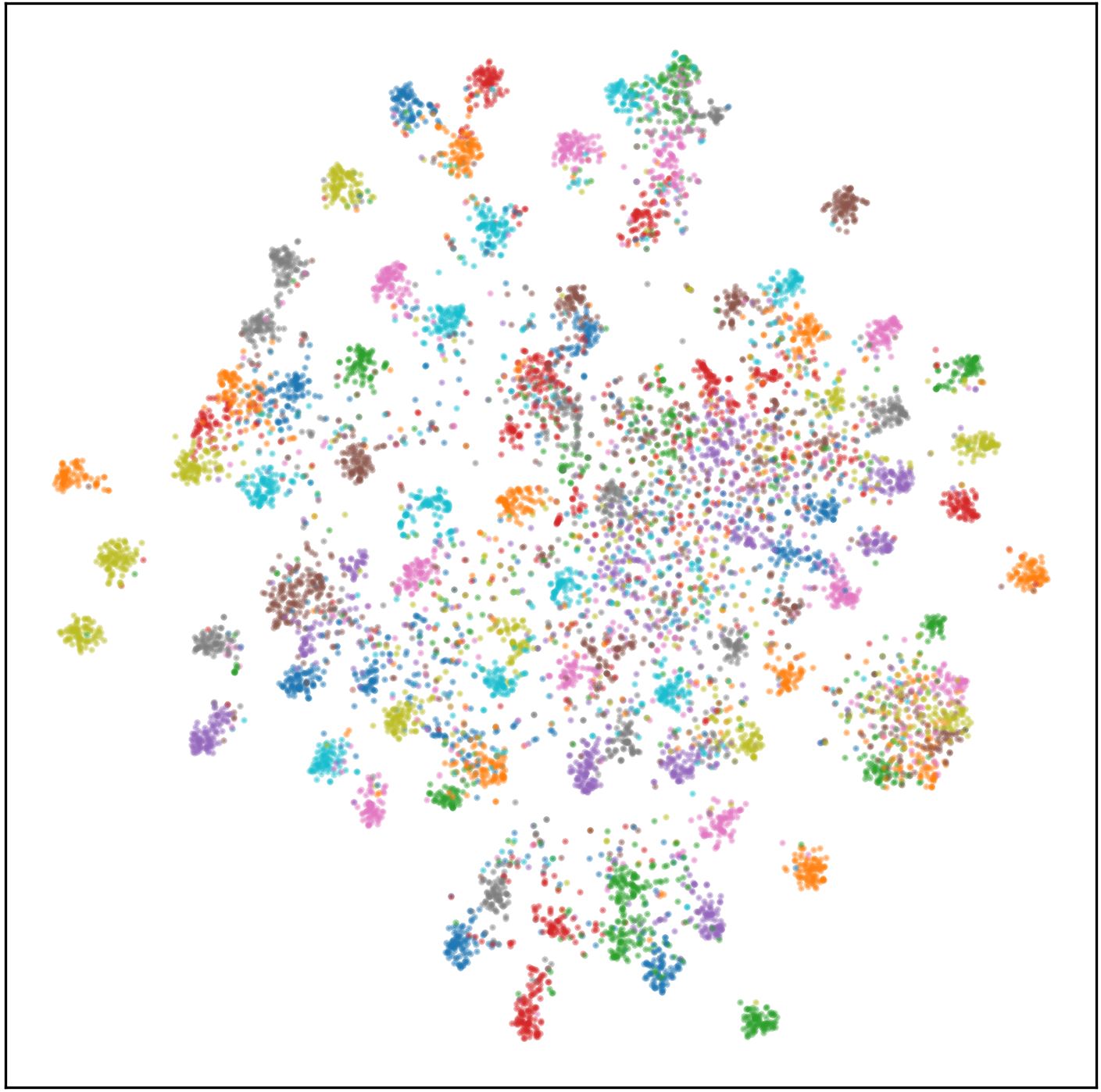}
		\caption{WTTM \cite{TTM}}
		\label{fig:ttm_resnet8x4_tsne}
	\end{subfigure}
	\begin{subfigure}{0.185\textwidth}
		\includegraphics[width=\textwidth]{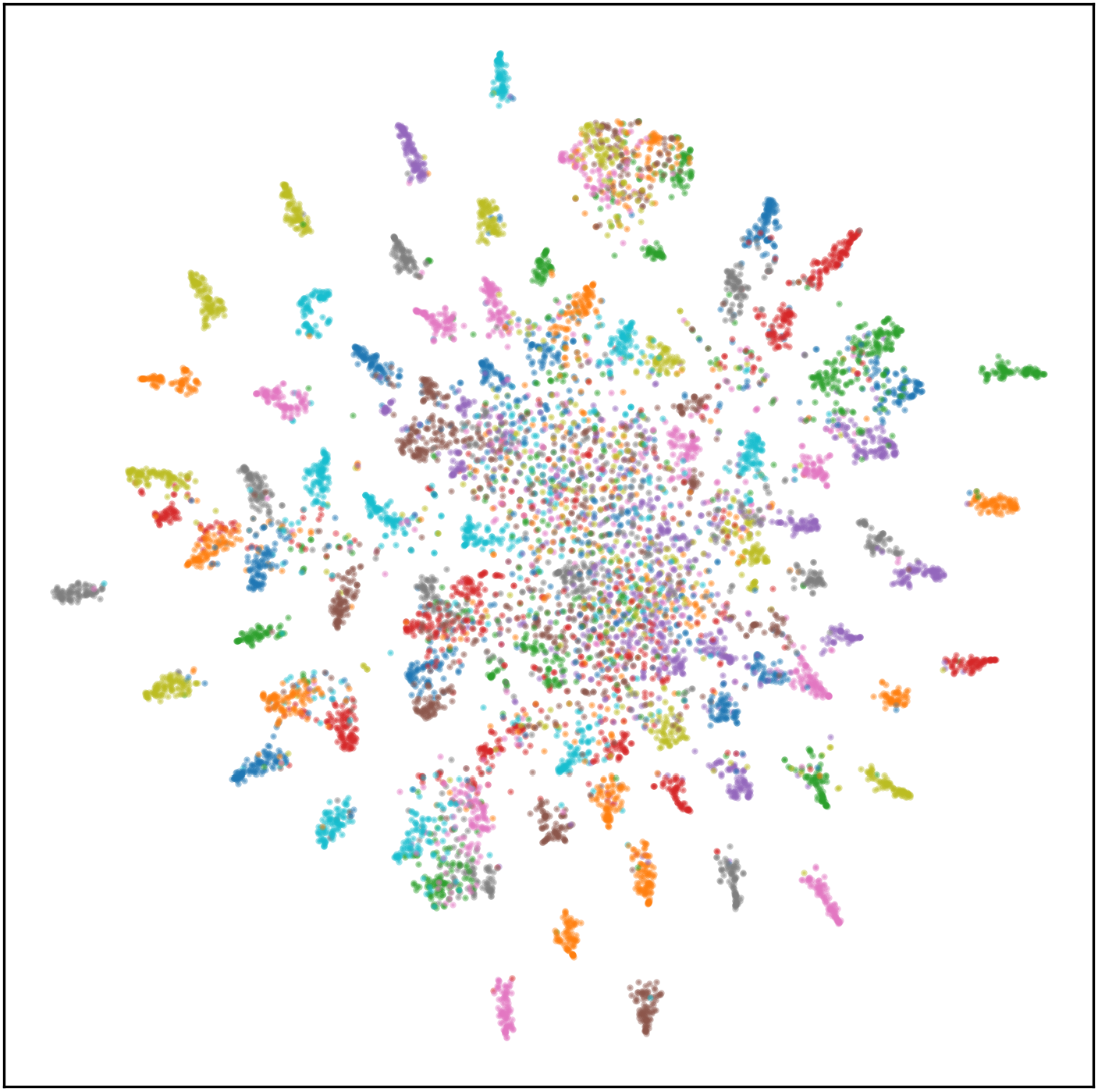}
		\caption{\textbf{Ours}}
		\label{fig:tsne_our}
	\end{subfigure}
	\begin{subfigure}{0.185\textwidth}
		\includegraphics[width=\textwidth]{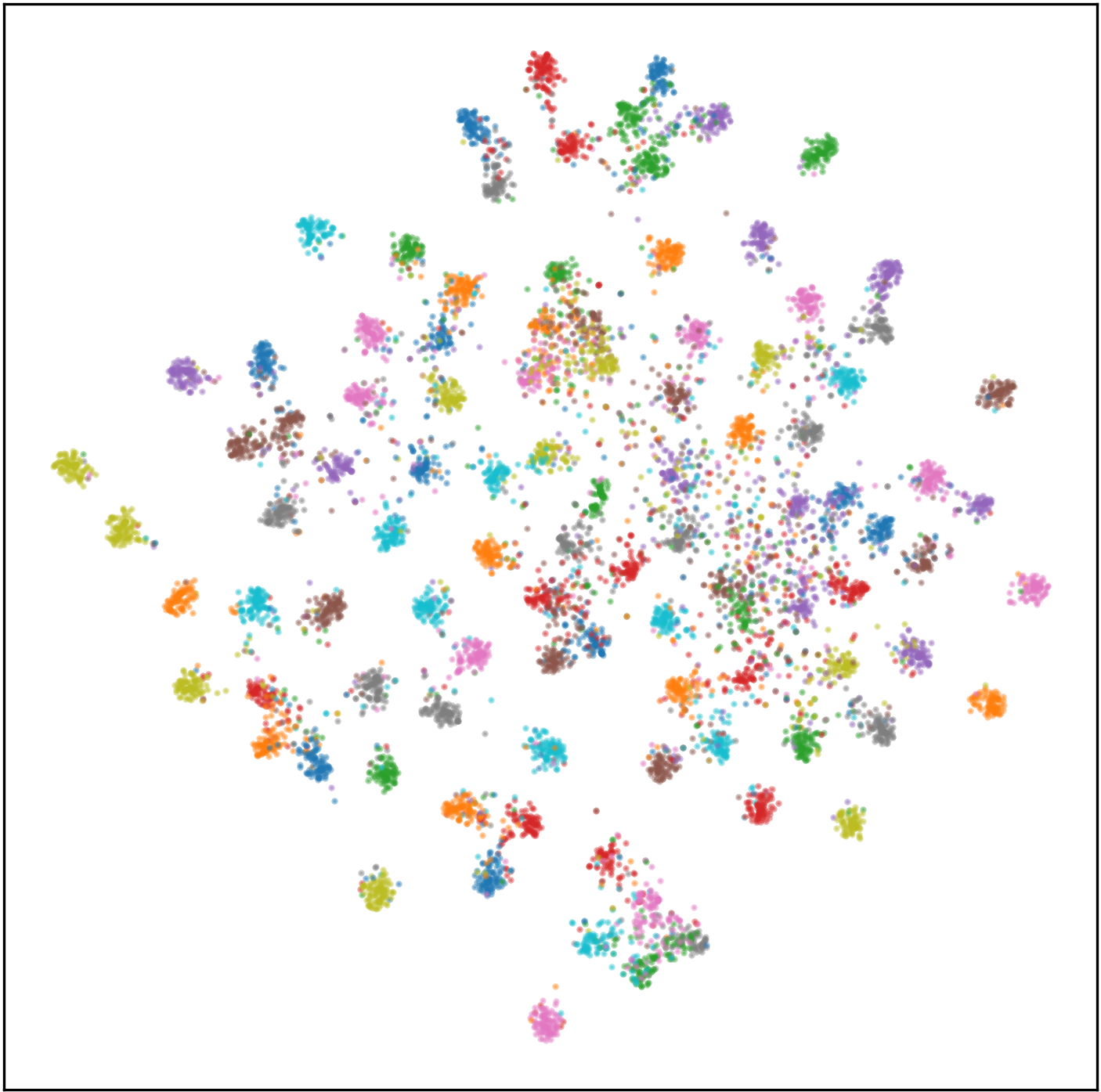}
		\caption{Teacher}
		\label{fig:tsne_teacher}
	\end{subfigure}
	\caption{The t-SNE \cite{tsne} visualization of features in a \textbf{homogeneous structure}. We set ResNet32×4 as the teacher model and ResNet8×4 as the student model. Our method achieves more separate classification results.}
	\label{fig:tsne_comparison}
\end{figure*}
\begin{figure*}[htbp]
	\centering
		\begin{subfigure}{0.185\textwidth}
		\includegraphics[width=\textwidth]{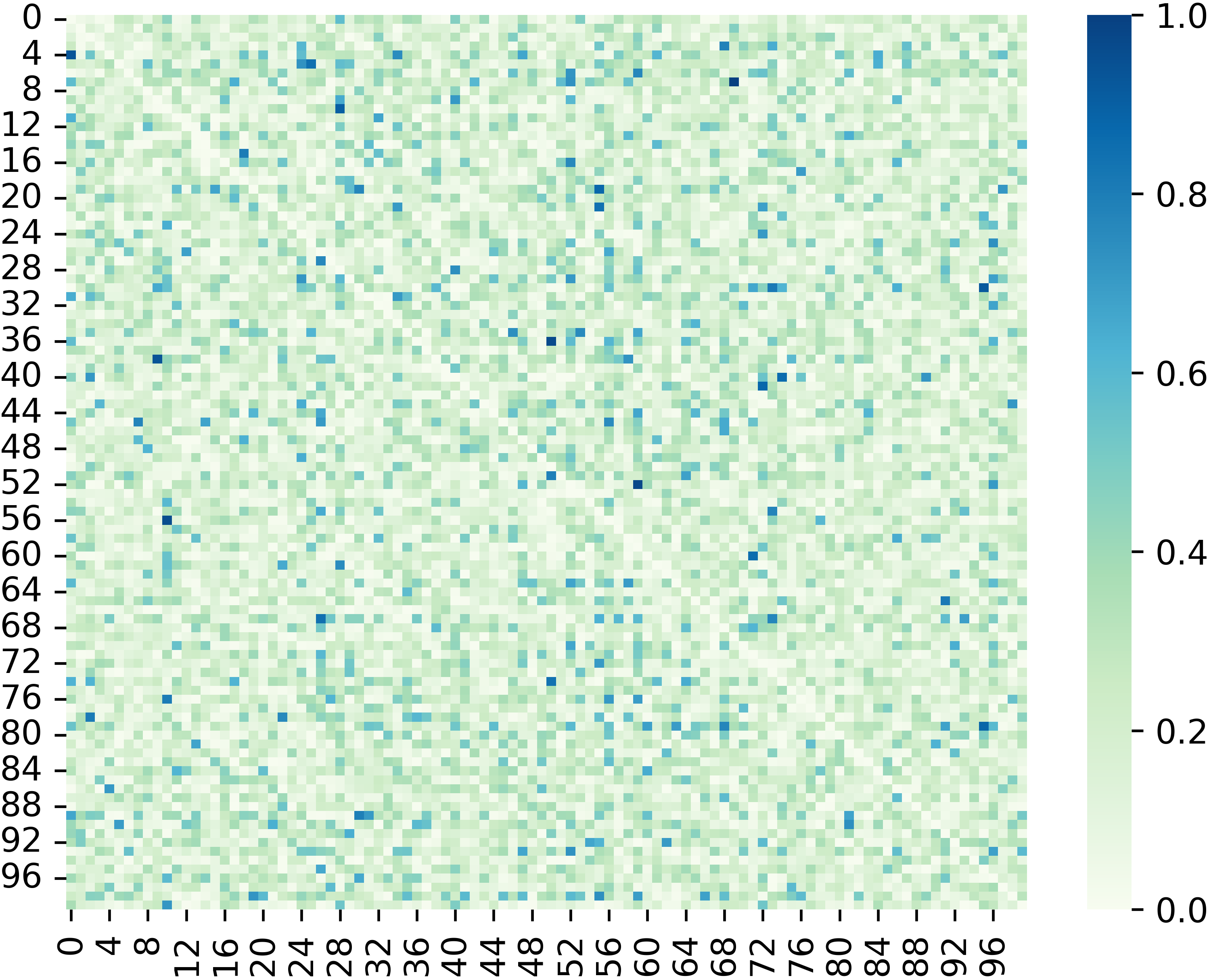}
		\caption{FCFD \cite{FCFD}}
		\label{fig:fcfd_correlation}
	\end{subfigure}
	\begin{subfigure}{0.185\textwidth}
		\includegraphics[width=\textwidth]{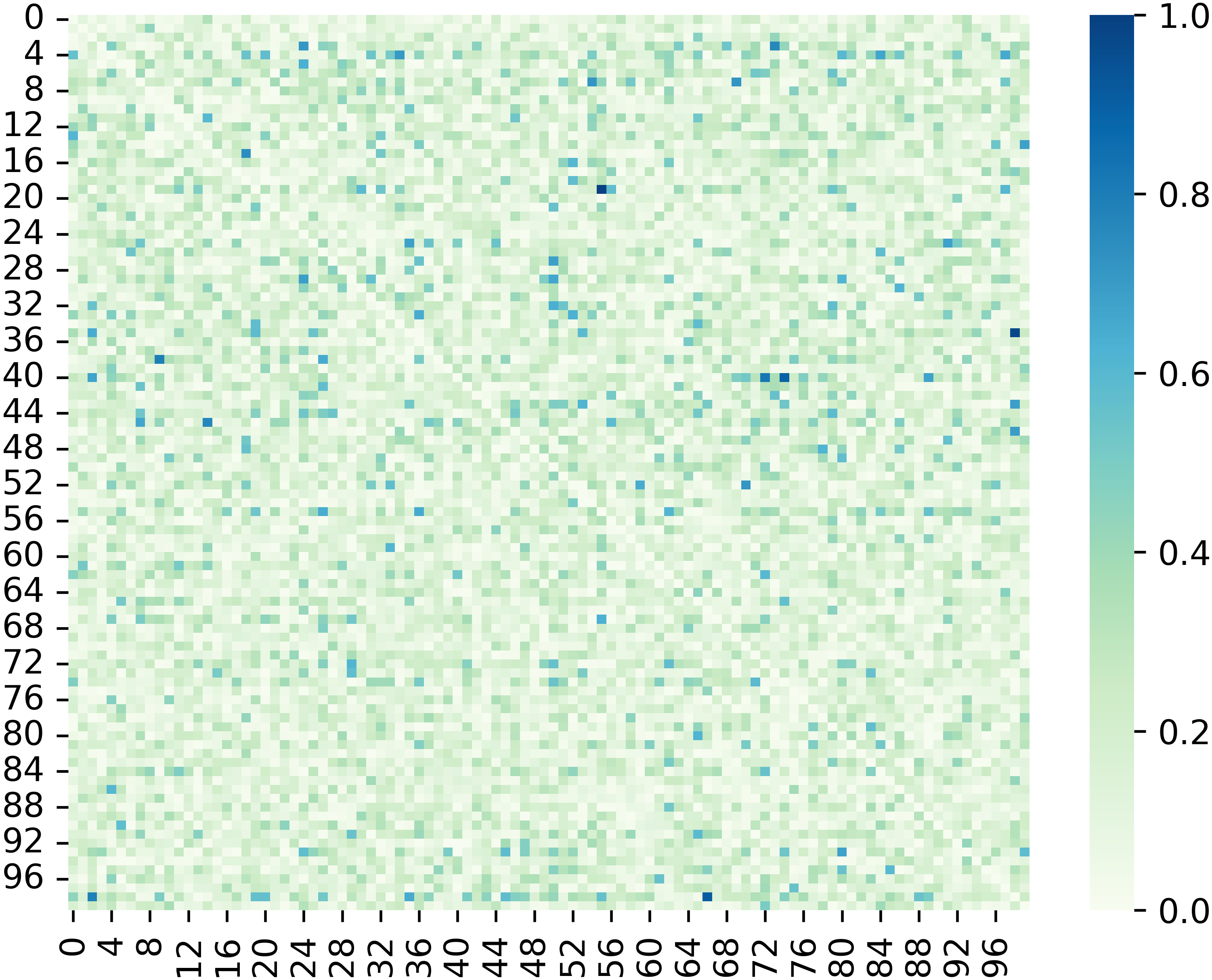}
		\caption{LSKD+DKD \cite{lskd}}
		\label{fig:lskd_correlation}
	\end{subfigure}
	\begin{subfigure}{0.185\textwidth}
		\includegraphics[width=\textwidth]{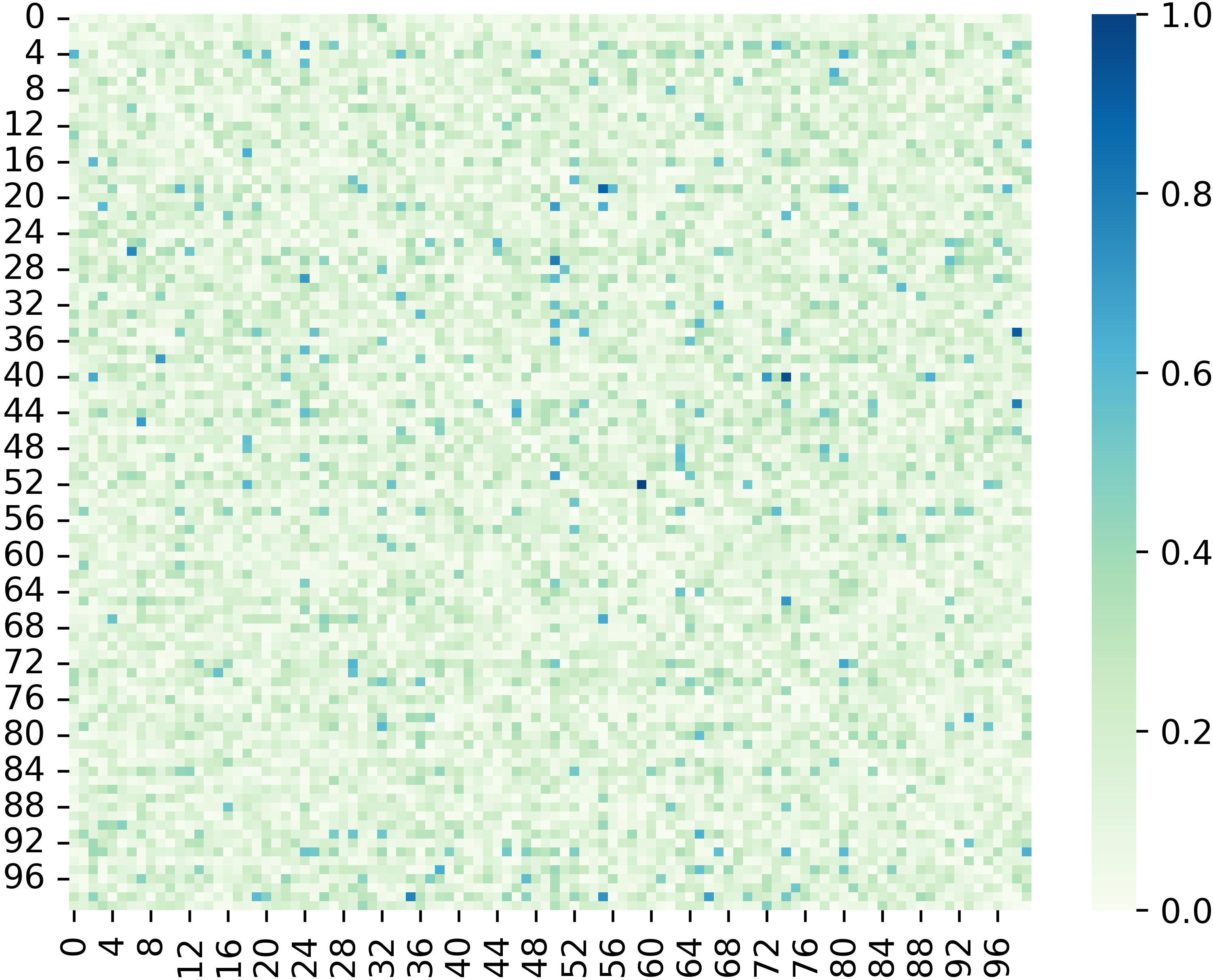}
		\caption{DOT+DKD \cite{dot}}
		\label{fig:dot_correlation}
	\end{subfigure}
	\begin{subfigure}{0.185\textwidth}
		\includegraphics[width=\textwidth]{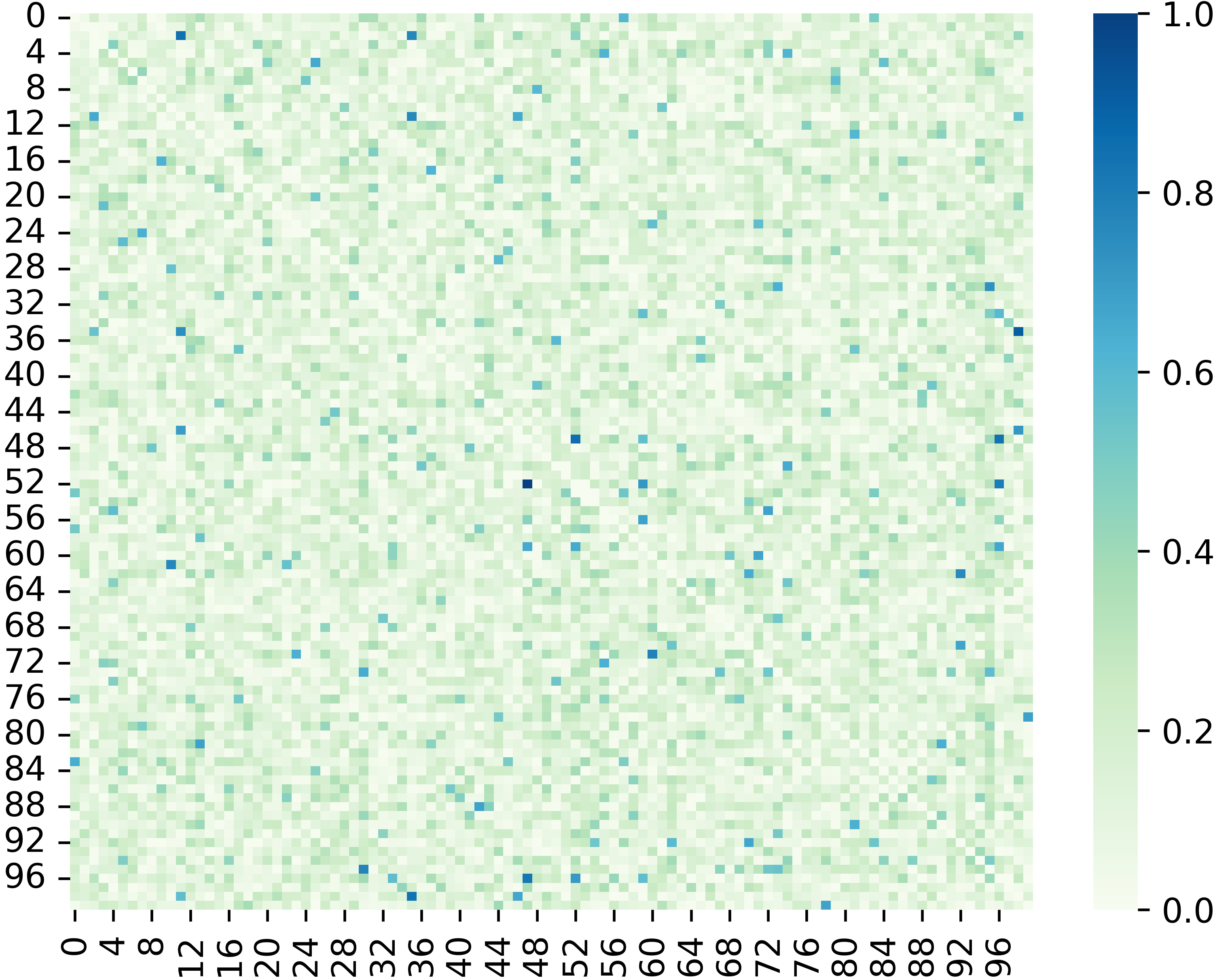}
		\caption{WTTM \cite{TTM}}
		\label{fig:wttm_correlation}
	\end{subfigure}
	\begin{subfigure}{0.185\textwidth}
		\includegraphics[width=\textwidth]{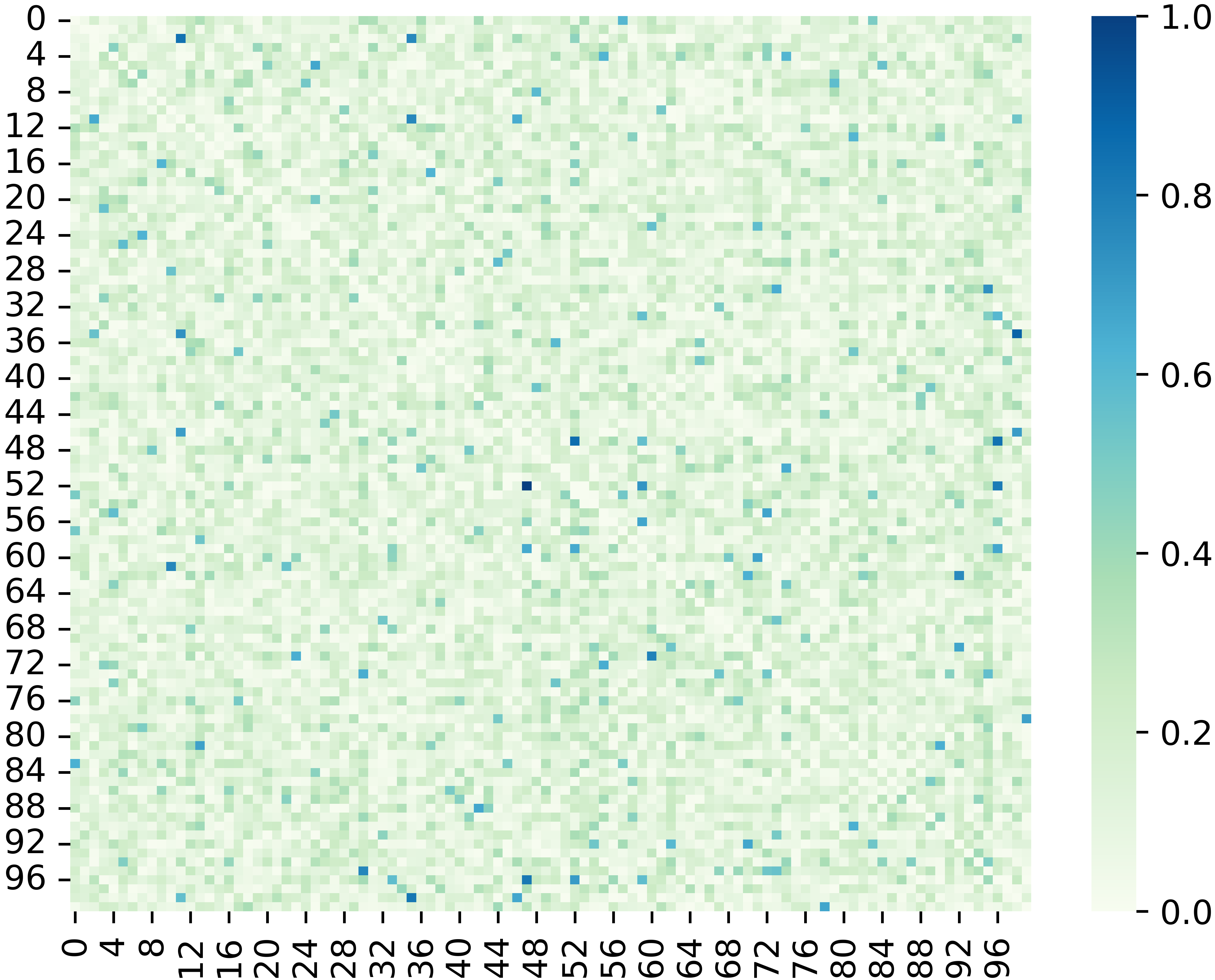}
		\caption{\textbf{Ours}}
		\label{fig:ours_correlation}
	\end{subfigure}
	\caption{Difference in correlation matrices between the student's and teacher's logits in a \textbf{heterogeneous structure}. We set WRN-40-2 as the teacher and ResNet8x4 as the student. The lighter the shade, the smaller the difference.}
	\label{fig:correlation_comparison}
\end{figure*}
\subsection{Extensions}
\label{extensions}
To achieve a more comprehensive understanding of MCLD, we analyze it from five perspectives. First, we evaluate the transferability of representations to assess MCLD's ability to generalize across different datasets. Second, we conduct more comprehensive experiments on the issue that ``bigger models are not always better teachers" and demonstrate how MCLD mitigates this problem through multi-perspective comparisons, even with large teacher-student discrepancies. Third, we provide visualizations that clearly validate the superiority of MCLD. Fourth, we compare the training efficiency of MCLD with state-of-the-art methods, highlighting how it balances model performance and computational cost. Finally, we present additional experiments, including the distillation of Vision Transformers and applications to other tasks such as object detection.\par
\noindent\textbf{Transferability of representations.} Knowledge distillation facilitates the transfer of unseen knowledge to new tasks or datasets not encountered during the original training. To evaluate the transferability of distilled representations, we use ResNet8x4 and MobileNetV2 as student models, which learn from ResNet32x4 and ResNet50, respectively, or are trained from scratch on CIFAR-100. For STL-10 \cite{stl-10} and Tiny-ImageNet \cite{tiny-imagenet} images (both resized to 32×32), the student model serves as a frozen feature extractor up to the layer preceding the logits. We then train a linear classifier for 10-class classification on STL-10 and 200-class classification on Tiny-ImageNet to assess the transfer of representations across datasets. The results in \cref{tab:transfer-res8x4} and \cref{tab:transfer-mv2} clearly demonstrate the strong transferability and generalization of features learned through our MCLD.\par
\begin{table}[htbp]
	\centering
	\caption{Transferability of representations from CIFAR-100 \cite{cifar100} to STL-10 \cite{stl-10} and Tiny-ImageNet \cite{tiny-imagenet} in a \textbf{homogeneous structure}. DOT is implemented with DOT+DKD.}
	\label{tab:transfer-res8x4}
	\resizebox{\columnwidth}{!}{%
		\begin{tabular}{@{}c|cccccc@{}}
			\toprule
			Method   & Res8x4 & DOT \cite{dot} & FCFD \cite{FCFD} & \textbf{Ours} & Res32x4 \\ \midrule
			STL-10   & 65.18 & 68.88 & \underline{71.26} & \textbf{71.40} & 66.59     \\
			TI       & 34.75 & 33.54 & \underline{36.89} & \textbf{37.47} & 29.65 \\ \bottomrule
		\end{tabular}%
	}
\end{table}
\begin{table}[htbp]
	\centering
	\caption{Transferability of representations from CIFAR-100 \cite{cifar100} to STL-10 \cite{stl-10} and Tiny-ImageNet \cite{tiny-imagenet} in a \textbf{heterogeneous structure}. LSKD is implemented with LSKD+DKD.}
	\label{tab:transfer-mv2}
	\resizebox{\columnwidth}{!}{%
		\begin{tabular}{@{}c|cccccc@{}}
			\toprule
			Method  & MN-V2 & LSKD \cite{lskd} & WTTM \cite{TTM} & \textbf{Ours} & ResNet50 \\ \midrule
			STL-10  & 65.61 & \underline{67.12} & 66.01 & \textbf{68.99} & 72.81 \\
			TI 	    & 28.08 & \underline{33.57} & 31.45 & \textbf{35.56} & 40.84 \\ \bottomrule
		\end{tabular}%
	}
\end{table}
\noindent\textbf{Bigger models are not always better teachers.} Stronger teacher models are generally expected to yield better results; however, they often fail to transfer valuable knowledge effectively and may even underperform compared to weaker teachers. Previous studies have attributed this phenomenon to the significant capability gap between larger teacher models and smaller student models \cite{dkd,lskd}. However, we argue that the core issue lies in the limited perspective from which logits have traditionally been viewed. Most existing logit distillation methods compute logits directly, which is inefficient and suboptimal. Teacher logits are highly semantic, and learning them directly prevents students from capturing much of the valuable underlying information. Our MCLD addresses this problem, as shown by the detailed comparisons in \cref{tab:stronger-teacher} and \cref{tab:stronger-diff-teacher}.\par
\begin{table}[htbp]
	\centering
	\caption{Results on CIFAR-100 \cite{cifar100} in a \textbf{homogeneous structure}. We set WRN-16-2 as the student and WRN series networks as the teachers.}
	\label{tab:stronger-teacher}
	\resizebox{\columnwidth}{!}{%
		\begin{tabular}{@{}c|cccccc@{}}
			\toprule
			Teacher  & WRN-28-2 & WRN-40-2 & WRN-16-4 & WRN-28-4 \\
			Acc      & 75.45 & 75.61 & 77.51  & 78.60  \\ \midrule
			DKD \cite{dkd}      & 75.92 & 76.24 & 76.00  & 76.45  \\
			FCFD \cite{FCFD}      & 74.80 & \underline{76.34} & 75.99  & \underline{76.56}  \\
			WTTM \cite{TTM}    & \underline{76.08} & 76.29 & \underline{76.64}  & 76.05  \\
			\textbf{Ours}     & \textbf{76.83} & \textbf{76.37} & \textbf{76.97}  & \textbf{77.40}  \\
			\bottomrule
		\end{tabular}%
	}
\end{table}
\begin{table}[htbp]
	\centering
	\caption{Results on CIFAR-100 \cite{cifar100} in a \textbf{heterogeneous structure}. We set WRN-16-2 as the student and used different networks as the teachers. The / symbol indicates an inability to reproduce results for the current model pair.}
	\label{tab:stronger-diff-teacher}
	\resizebox{\columnwidth}{!}{%
		\begin{tabular}{@{}c|cccccc@{}}
			\toprule
			Teacher  & Vgg13 & WRN-16-4 & ResNet50 & ResNet32x4 \\
			Acc      & 74.64 & 77.51 & 79.34  & 79.42  \\ \midrule
			DKD \cite{dkd}     & 75.45 & 76.00 & \underline{76.60}  & 75.70  \\
			FCFD \cite{FCFD}    & / & 75.99 & /  & \underline{76.73}  \\
			WTTM \cite{TTM}    & \underline{75.63} & \underline{76.64} & 76.17  & 76.04  \\
			\textbf{Ours}     & \textbf{76.26} & \textbf{76.97} & \textbf{77.00}  & \textbf{77.44}  \\
			\bottomrule
		\end{tabular}%
	}
\end{table}
\noindent\textbf{Visualization.} We present visualization results from two perspectives on CIFAR-100 \cite{cifar100}. (1) The t-SNE \cite{tsne} results (\cref{fig:tsne_comparison}) using ResNet32x4 as the teacher and ResNet8x4 as the student show that the representations learned through MCLD are more distinguishable than those learned through other distillation methods. (2) Additionally, we visualize the absolute difference between the student's and teacher's logit correlation matrices (\cref{fig:correlation_comparison}), using WRN-40-2 as the teacher and ResNet8x4 as the student.\par
\noindent\textbf{Training Efficiency.} We compare the training time of logit distillation and feature distillation methods, highlighting the superior efficiency of MCLD. As shown in \cref{fig:kd-comparison}, MCLD achieves a trade-off between performance and training time. Due to limitation of page, detailed training time and training accuracy of most methods are attached in supplements.\par
\noindent\textbf{Vision Transformers and Detection} More experiments on distilling Vision Transformers \cite{vit-at,lg,autokd,tinyvit,deti,t2t,pit,pvt}, and more datasets (i.e., MS-COCO) \cite{coco}, are attached in the supplements. Notably, our MCLD consistently achieves superior performance across most vision transformers.
\section{Conclusion}
\label{sec:conclusion}
In this study, we revisit the computational mechanisms of traditional logit distillation methods and propose a novel approach (\textbf{MCLD}) that learns logits. While logits inherently encapsulate rich semantic information, traditional logit distillation methods fail to exploit the semantic properties effectively. To address this issue, we introduce contrastive learning to train logits, allowing the model to discern subtle distinctions more effectively. Furthermore, our multi-perspective framework optimizes the utilization of logits, providing comprehensive and structured guidance during training. Notably, our MCLD achieves better performance with less training time, more effectively addressing the issue ``bigger models are not always better teachers" while also exhibiting superior transfer learning capabilities. Extensive experiments validate that MCLD consistently outperforms most state-of-the-art logit and feature distillation methods, highlighting its effectiveness and superiority. We will release code, training weights at the appropriate time.\newpage
{
    \small
    \bibliographystyle{ieeenat_fullname}
    \bibliography{main}
}
\end{document}